\RequirePackage{fix-cm}
\documentclass[twocolumn]{svjour3}          
\smartqed  
\usepackage[hyphens]{url}
\usepackage[utf8]{inputenc}
\usepackage{graphicx,subfig,diagbox,textcomp}
\usepackage{multirow}
\usepackage{color}

\usepackage{gensymb}
\usepackage{amsmath}
\usepackage{amsfonts}
\usepackage{amssymb}
\usepackage{hyperref}
\hypersetup{breaklinks=true}
\usepackage[english]{babel}
\usepackage{soul}

\usepackage[table]{xcolor}
\usepackage{booktabs}

\usepackage{mathptmx} 
\journalname{Multimedia Tools and Applications}
\newcommand{\minisection}[1]{\vspace{0.04in} \noindent {\bf #1}\ \ }
\newcommand{\tabitem}{~~\llap{\textbullet}~~}
\begin{document}

\title{Review on Computer Vision Techniques in Emergency Situations
\thanks{This work was partially supported by the Spanish Grants TIN2016-75404-P AEI/FEDER, UE, TIN2014-52072-P, TIN2013-42795-P and the European Commission H2020 I-REACT project no. 700256. Laura Lopez-Fuentes benefits from the NAERINGSPHD fellowship of the Norwegian Research Council under the collaboration agreement Ref.3114 with the UIB. We thank the NVIDIA Corporation for support in the form of GPU hardware.}
}


\author{Laura Lopez-Fuentes \and
Joost van de Weijer  \and
        Manuel~González-Hidalgo \and
        Harald Skinnemoen \and
        Andrew D. Bagdanov         
}


\institute{L. Lopez-Fuentes\at
              TM-RCS Working Group, AnsuR Technologies AS, Fornebu. Norway.\\ University of the Balearic Islands, Dept of Mathematics and Computer Science, Crta. Valldemossa, km 7.5, E-07122 Palma, Spain.\\              
Computer Vision Center. Universitat Autònoma de Barcelona. 08193-Barcelona. Spain.
              \email{l.lopez@uib.es}           
           \and
           J. van de Weijer \at
              Computer Vision Center. Universitat Autònoma de Barcelona. 08193-Barcelona. Spain.
              \email{joost@cvc.uab.es}
              \and M. González-Hidalgo \at University of the Balearic Islands, Dept of Mathematics and Computer Science, Crta. Valldemossa, km 7.5, E-07122 Palma, Spain.
              \email{manuel.gonzalez@uib.es}
              \and H. Skinnemoen \at
              TM-RCS Working Group, AnsuR Technologies AS, Fornebu. Norway.
              \email{harald@ansur.no}
              \and A. D. Bagdanov \at 
              University of Florence, DINFO, Via Santa Marta 3, 50139 Florence, Italy.
              \email{andrew.bagdanov@unifi.it}         
}

\date{Received: date / Accepted: date}

\maketitle

\begin{abstract}
In emergency situations, actions that save lives and limit the impact of hazards are crucial. In order to act, situational awareness is needed to decide what to do. Geolocalized photos and video of the situations as they evolve can be crucial in better understanding them and making decisions faster. Cameras are almost everywhere these days, either in terms of smartphones, installed CCTV cameras, UAVs or others. However, this poses challenges in big data and information overflow. Moreover, most of the time there are no disasters at any given location, so humans aiming to detect sudden situations may not be as alert as needed at any point in time. Consequently, computer vision tools can be an excellent decision support. The number of emergencies where computer vision tools has been considered or used is very wide, and there is a great overlap across related emergency research. Researchers tend to focus on state-of-the-art systems that cover the same emergency as they are studying, obviating important research in other fields. In order to unveil this overlap, the survey is divided along four main axes: the types of emergencies that have been studied in computer vision, the objective that the algorithms can address, the type of hardware needed and the algorithms used. Therefore, this review provides a broad overview of the progress of computer vision covering all sorts of emergencies. 
\end{abstract}

\section{Introduction}
\label{intro}
Emergencies are a major cause of both human and economic loss. They
vary in scale and magnitude, from small traffic accidents involving
few people, to full-scale natural disasters which can devastate
countries and harm thousands of people. \emph{Emergency management}
has become very important to reduce the impact of emergencies.
As a consequence, using modern technology to implement innovative
solutions to prevent, mitigate, and study emergencies is an active
field of research.

\begin{figure*}
\begin{center}
\setlength{\tabcolsep}{0.1em}
\begin{tabular}{cccc}
\subfloat{\includegraphics[width = 0.25\linewidth]{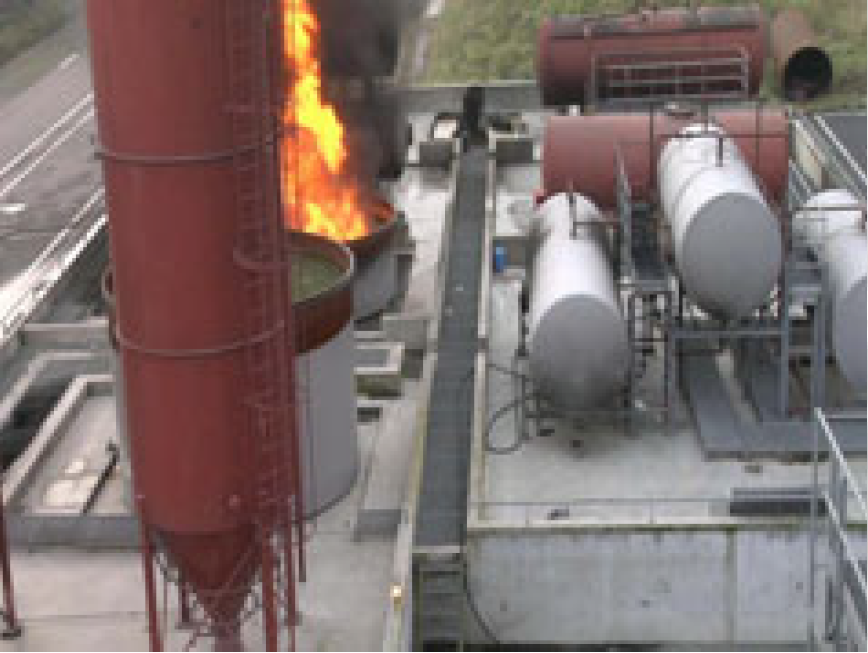}}&
\subfloat{\includegraphics[width = 0.25\linewidth]{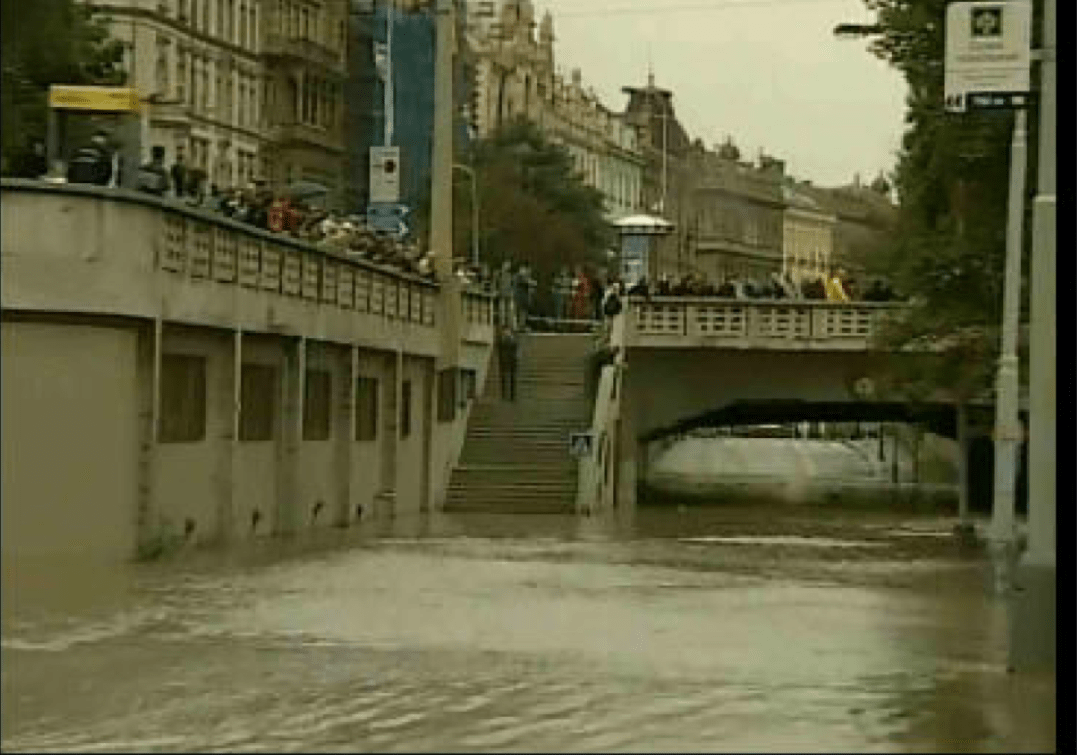}} &
\subfloat{\includegraphics[width = 0.25\linewidth]{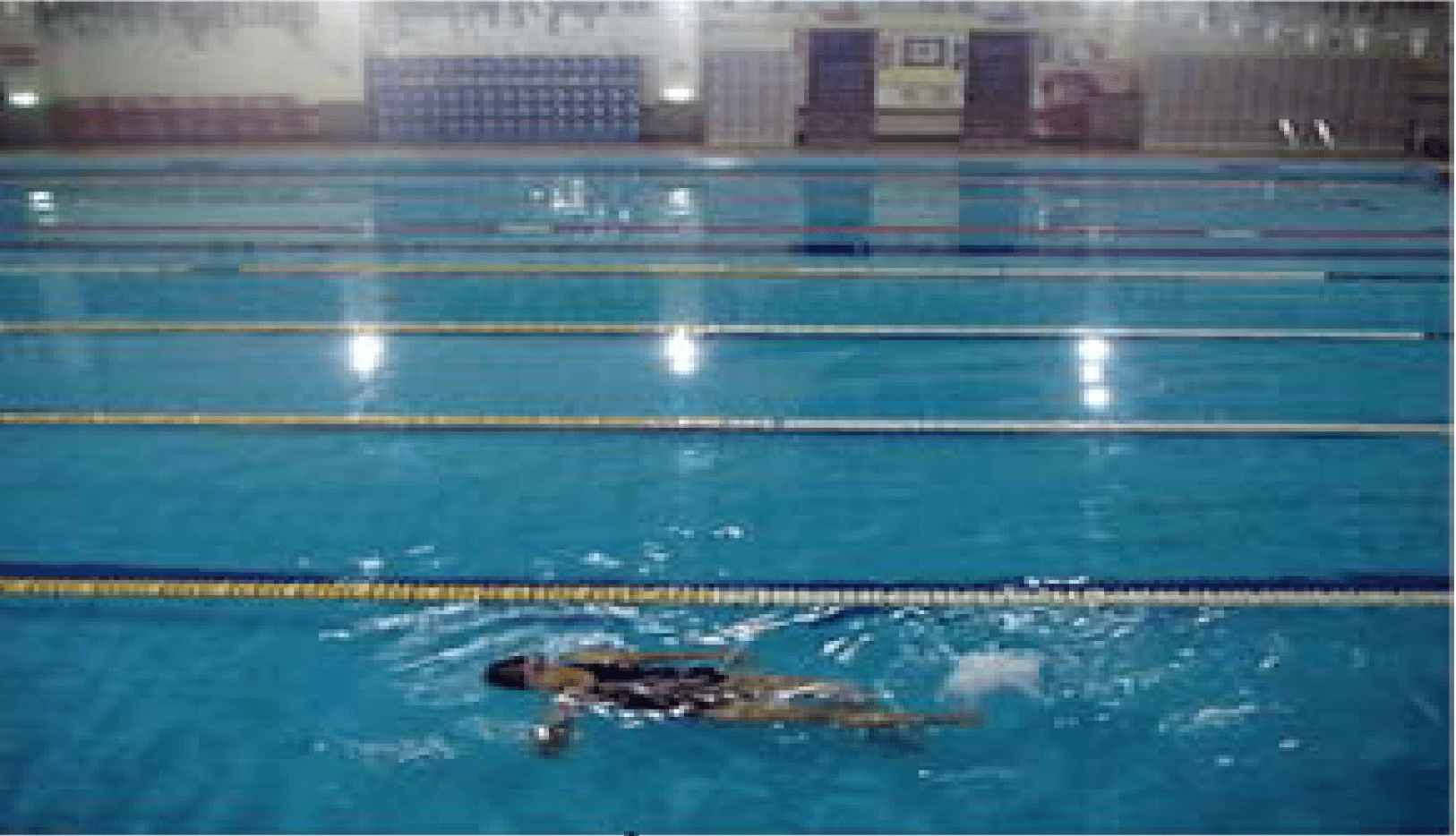}} &
\subfloat{\includegraphics[width = 0.25\linewidth]{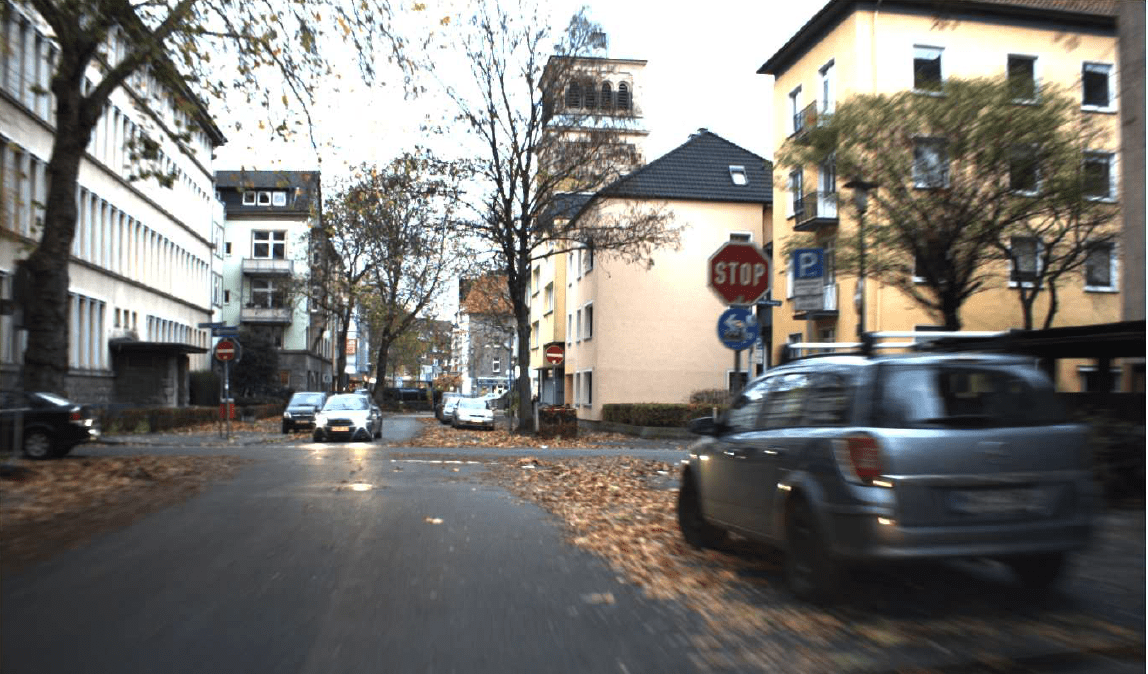}}
\end{tabular}
\caption{Example frames from videos that could potentially be
  recorded to detect different emergencies: a factory being
  monitored to for rapid fire detection \cite{van2010fire}, a street
  monitored to detect floods \cite{borges2008probabilistic}, a
  surveillance camera on a pool to detect drowning persons,
  \cite{chen2010hidden} and a vehicle with on board camera for driver
  assistance \cite{houben2013detection}.}
\label{fig:examples_surveillance_cameras}
\end{center}
\end{figure*}

The presence of digital cameras has grown explosively over the last
two decades. Camera systems perform real-time recording of visual
information in hotspots where an accident or emergency is likely to
occur. Examples of such systems are cameras on the road, national
parks, inside banks, shops, airports, metros, streets and swimming
pools. In Figure \ref{fig:examples_surveillance_cameras} we give
some examples of frames from videos that could potentially be recorded
by some of these monitoring systems. It is estimated that as of 2014
there were over 245 million active, operational and professionally
installed video cameras around the
world~\cite{IHS2015surveillance}. Mounting these video cameras has
become very cheap but there are insufficient human resources to
observe their output. Therefore, most of the data is being processed
\emph{after} an emergency has already occurred, thus losing the
benefit of having a real-time monitoring tool in the first place.

The problem of real-time processing of the visual output from
 monitoring systems in potential emergency scenarios
has become a hot topic in the field of computer vision. As we will
show in this review, computer vision techniques have been used to
assist in various stages of emergency management: from prevention,
detection, assistance of the response, to the understanding of
emergencies. A myriad of algorithms has been proposed to prevent
events that could evolve into disaster and emergency
situations. Others have focused on fast response after an emergency
has already occurred. Yet other algorithms focus on assisting during
the response to an emergency situation. Finally, several methods focus
on improving our understanding of emergencies with the aim of
predicting the likelihood of an emergency or dangerous situation
occurring.

This review provides a broad overview of the progress of computer
vision covering all sorts of emergencies. The review is organized
in four main axes along which researchers working on the topic of
computer vision for emergency situations have organized their
investigations. The first axis relates to the \emph{type of emergency}
being addressed, and the most important division along this axis is
the distinction between \emph{natural} and \emph{man-made}
emergencies. The second axis determines the objective of a system: if
it is going to detect risk factors to prevent the emergency, help
detecting the emergency once it has occurred, provide information to
help emergency responders to react to the emergency, or model the
emergency to extract valuable information for the future. The third
axis determines where the visual sensors should be placed, and
also what type of sensors are most useful for the task. The last axis determines which
algorithms will be used to assist in analyzing the data. This
typically involves a choice of computer vision algorithms combined
with machine learning techniques.

When organizing emergency situations along these axes, our review shows that there is significant overlap across emergency research areas, especially in the sensory data and algorithms used. For this reason this review is useful for researchers in emergency topics, allowing them to more quickly find relevant works. Furthermore, due to its general nature, this article aims to increase the flow of ideas among the various emergency research fields.



Since it is such an active field of research, there are several
reviews on computer vision algorithms in emergency situations. However
these reviews focus on a single, specific topic. In the field of fire
detection, a recent survey was published summarizing the different
computer vision algorithms proposed in the literature to detect fire
in video~\cite{ccetin2013video}. Also Gade et
al.~\cite{gade2014thermal} published a survey on applications of
thermal cameras in which fire detection is also considered. Another
related research direction is human activity recognition and detection
of abnormal human actions. This field has many applications to video
surveillance, as shown in a recent review on this
topic~\cite{ke2013review}. Work in the area of human activity
recognition has focused on the specific topic of falling person
detection. Mubashir et al.~\cite{gade2014thermal} describe the three
main approaches to falling person detection, one of them being
computer vision. Another important application area of emergency
management is traffic. The number of vehicles on roads has
significantly increased in recent years, and simultaneously the number
of traffic accidents has also increased. Therefore a large number of
road safety infrastructure and control systems have been
developed. Three main reviews have been recently published on the
topic of road vehicles monitoring and accident detection using
computer
vision~\cite{buch2011review,kanistras2015survey,kovacic2013computer}. As
discussed here, there are several review papers about different
specific emergency situations, however there is no \emph{one} work
reviewing computer vision in emergency situations considering many
types of emergency situations. By doing so, we aspire to draw a wider
picture of the field, and to promote cross-emergency dissemination of
research ideas.

Computer vision is just one of several scientific fields which aims to improve emergency management. Before going into depth on computer vision techniques for emergency management, we briefly provide some pointers to other research fields for interested readers. Emergency management is a hot topic in the field of Internet of Things (IoT) \cite{maalel2013reliability,yang2013internet}. The goal of the IoT is to create an environment in which information from any object connected to the network be efficiently shared with others in real-time. The authors of \cite{yang2013internet} provide an excellent study on how IoT technology can enhance emergence rescue operations. Robotics research is another field which already is changing emergency management. Robots are being studied for handling hazardous situations with radiation \cite{nagatani2013emergency}, for assisting crowd evacuation \cite{sakour2017robot}, and for search and rescue operations \cite{shah2004survey}. Finally, we mention reviews on human centered sensing for emergency evacuation \cite{radianti2013crowd} and research on the impact of social media on emergency management. The main advantage of computer vision compared to these other research fields is the omni-presence of cameras in modern society, both in the public space and mobile phones of users. However, for some problems computer vision might currently be still less accurate than other techniques.

\begin{figure*}[t!]
\centerline{\includegraphics[width = 0.8\linewidth]{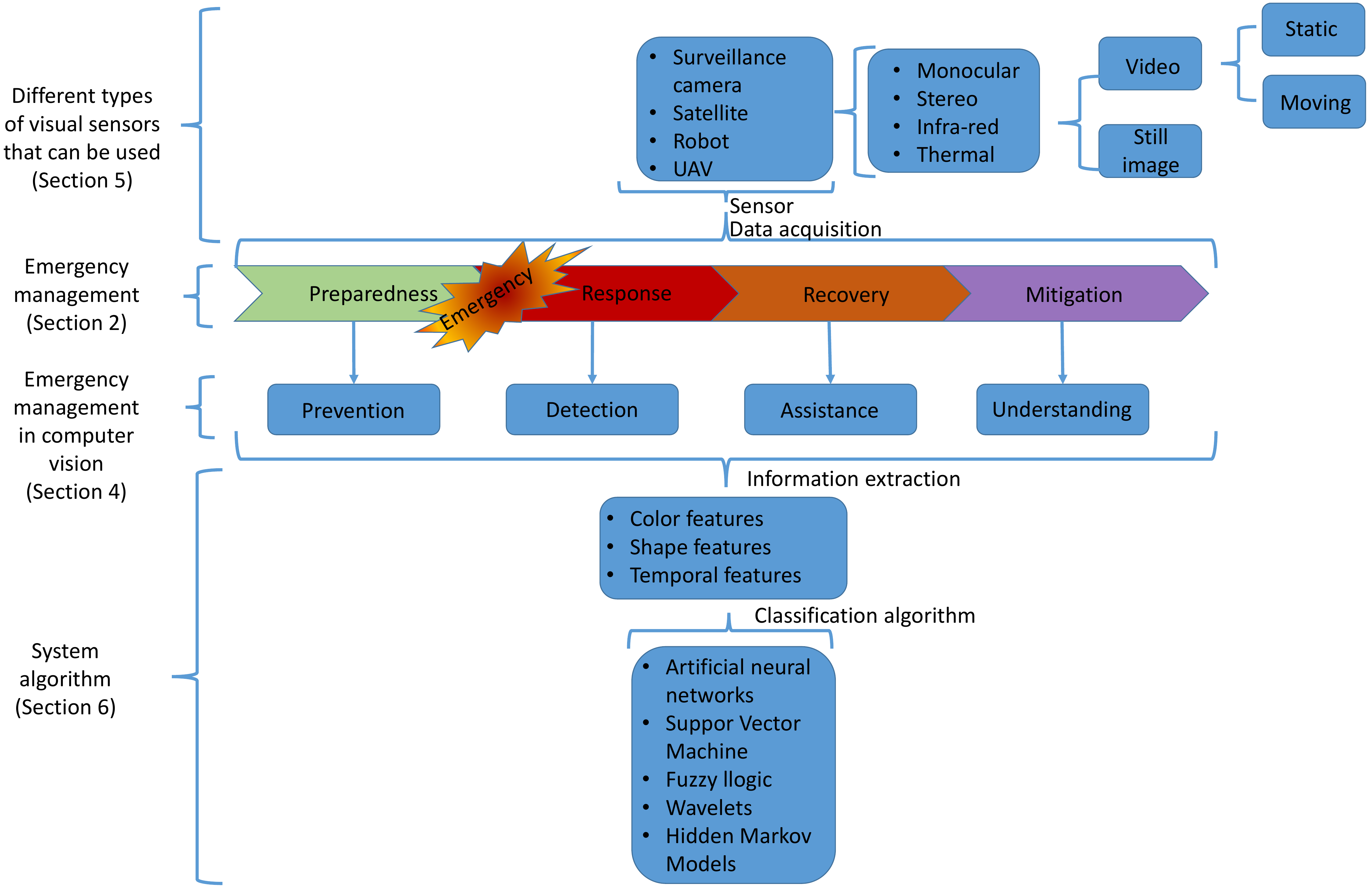}}
\caption{A graphical overview of the organization of this survey.}\label{fig:graph}
\end{figure*}

\section{Organization of this Review}
\label{sec:organization}
Before we enter into the organization of the review
it is important to define what we consider as an \emph{emergency} and,
therefore, what situations and events will be encompassed in this
definition. We consider as an emergency any sudden or unexpected
situation that meets at least one of the following conditions:
\begin{itemize}
\item it is a situation that represents an imminent risk of health or
  life to a person or group of persons or damage to properties or
  environment;
\item it is a situation that has already caused loss of health or life
  to a person or group of persons or damage to properties or
  environment; or
\item it is a situation with high probability of scaling, putting in
  risk the health or life of a person or group of persons or damage to
  properties or environment.
\end{itemize}
An emergency situation, therefore, requires situational awareness,
rapid attention, and remedial action to prevent degradation of the
situation.  In some cases emergency situations can be prevented by
avoiding and minimizing risks. Also, in some situations, the
occurrence of an emergency situation can help to understand risk
factors and improve the prevention mechanisms. Taking this definition in consideration, any type of uncontrolled fire
would be considered as an emergency situation as it is an unexpected
situation that could represent a risk for persons, the environment,
and property. A controlled demolition of a building would not be
considered an emergency situation, as by definition if it is
controlled it is not an unexpected situation any more. However, if
someone is injured or something unforeseen was damaged during the
demolition, that would then be considered an emergency situation.

\begin{table*}
\begin{center}
\begin{tabular}{ccc}
\toprule
{\bf Generic Emergency Group} & {\bf Emergency sub-group} & {\bf Emergency Type}
\\ \midrule
Natural emergencies & &
\begin{tabular}{c}
Fire \\  Flood \\  Drought \\ Earthquake \\ Hurricane/Tornado \\ Landslide/Avalanche
\end{tabular} \\
\midrule
\multirow{6}{*}{Man-made emergencies} & \multicolumn{1}{c}{\multirow{3}{*}{Single person}}& Falling person\\  \multicolumn{1}{c}{}&& Drowning \\  \multicolumn{1}{c}{}&& Injured civilians \\ \cmidrule(r){2-3} \multicolumn{1}{c}{}&
\multirow{3}{*}{Multiple person} & Road accident\\   \multicolumn{1}{c}{}&& Crowd related \\   \multicolumn{1}{c}{}&& Weapon threaten \\
\bottomrule
\end{tabular}
\end{center}
\caption{Types of emergencies and their classification.}
\label{tab:types of emergencies}
\end{table*}

After having defined what we consider an emergency situation
we briefly explain the organization of this review. We
begin in Section~\ref{sec:Definition} with an overview and
classification into subcategories of emergency situations. Here, we
briefly introduce the three axes along which we will organize the
existing literature on computer vision for emergency situations, which corresponds to Sections 4 to 6. The
organization is also summarized in Figure~\ref{fig:graph}.

\minisection{The Monitoring Objective Axis
  (Section~\ref{sec:objective}):} the management of an emergency has a
life cycle which can be divided into four main phases~\cite{petak1985emergency}:
\begin{itemize}
\item \textbf{Preparedness}: during this phase there are risk factors
  that could trigger an emergency at some point. In order to reduce
  the chance of an emergency happening, it is important to estimate
  these factors and avoid or minimize them as much as possible. Not
  all emergencies can be prevented, but in some cases preventive
  measures can help to reduce the impact of a disaster (like teaching
  people how to behave during a
  hurricane). 
\item \textbf{Response}: response activities are performed after the
  emergency has occurred. At this point, it is very important to
  detect and localize the emergency and to maintain situational
  awareness to have a rapid and efficient reaction. Often the minutes  immediately after the occurrence of an emergency can be
  crucial to mitigating its effects.
\item \textbf{Recovery}: recovery activities take place until
  normality is recovered. In this phase actions must be taken to save
  lives, prevent further damaging or escalation of the situation. We
  consider that normality has been recovered when the immediate threat
  to health, life, properties or environment has subsided.
\item \textbf{Mitigation}: this phase starts when normality has been
  re-established. During this phase it is important to study and
  understand the emergency, how it could have been avoided or
  what responses could have helped to mitigate it.
\end{itemize}

\minisection{The Sensors and Acquisition System Axis
  (Section~\ref{sec:acquisition}):} An emergency has four main management phases: prevention, detection, response and understanding of an emergency. For a computer vision algorithm to help in one or more of these phases, it is important to have visual sensors
to record the needed information. There are many different types of
devices for this, as also many types of devices in which they may be
installed depending on the type of data being captured. Some of the
most common are: fixed monitoring cameras, satellites, robots and
Unmanned Aerial Vehicles (UAVs). These devices can capture different
types of visual data, such as stereo or monocular images, infrared or
thermal. Depending on the rate of data acquisition this data can be in
the form of video or still images, and the video can be either from a
static or moving source.

\minisection{The Algorithmic Axis (Section~\ref{sec:algorithms}):} The
visual data collected by the sensor is then subject to a feature
extraction process. Finally the features extracted are used as an
input to a modeling or classification algorithm. In
Figure~\ref{fig:graph} the main feature extraction and classification
algorithms are listed.



\section{Types of Emergencies}
\label{sec:Definition}

There are many types of emergencies and a great deal of effort has
been invested in classifying and defining
them~\cite{abdallah2000public,below2009disaster}. However there are
many emergencies that due to their nature have not been studied in
computer vision, such as famine or pollution. These types of
emergencies do not have clear visual consequences that can be easily
detected by visual sensors. In this paper we focus only on emergencies
relevant in the computer vision field. In Table~\ref{tab:types of
  emergencies} we classify the types of emergencies that we consider
in this survey. We distinguish two generic emergency groups: natural
emergencies and man-made emergencies. Natural emergencies are divided
into three emergency types: fire, flood and drought. Emergencies caused by humans are divided into two
sub-types depending on the scope: emergencies that cause dangers to
multiple persons and emergencies that cause dangers to a single
person.

\subsection{Natural emergencies}
Natural emergencies cover any adverse event resulting from natural
processes of the earth. Natural emergencies can result in a loss of
lives and can have huge economic costs. The severity of the emergency
can be measured by the losses and the ability of the population to
recuperate. Although natural emergencies are external to humans and
therefore can not be prevented, society has been increasingly able to
reduce the impact of these natural events by building systems for
early detection or assistance during the event. Among these systems
built to detect or assist natural emergencies, computer vision has
played an important role in emergencies related to fire, flood and
drought.

\minisection{Fire:} electric fire detectors have been integrated into
buildings since the late 1980s. These devices detect smoke, which is a
typical fire indicator. Currently, two major types of electric smoke
detector devices are used~\cite{reisinger1980smoke}: ionization smoke
detectors and photoelectric smoke detectors. Although the devices
achieve very good performance in closed environments, their
reliability decreases in open spaces because they rely on smoke
particles arriving to the sensor. For this reason, fire detection in
monitoring cameras has become an important area of research in
computer vision. The use of video as input in fire detection makes it
possible to use in large and open spaces. Hence, most computer vision
algorithms that study fire detection focus mainly on open
spaces~\cite{martinez2008computer,gunay2010fire}. However, some
articles~\cite{ko2010early,truong2012fire} also study fire detection
in closed spaces, claiming to have several advantages over
conventional smoke detectors: visual fire detectors can trigger the
alarm before conventional electric devices, as they do not rely on the
position of fire particles, and they can give information about the
position and size of the fire.

\minisection{Flood:} flood results from water escaping its usual
boundaries in rivers, lakes, oceans or man-made canals due to an
accumulation of rainwater on saturated ground, a rise in the sea level
or a tsunami. As a result, the surrounding lands are inundated. Floods
can result in loss of life and can also have huge economic impact by
damaging agricultural land and residential areas. Flood detection or
modeling algorithms are still at a very early stage. In computer
vision, there are two main trends: algorithms based on still cameras
monitoring flood-prone areas~\cite{lai2007real} and systems that
process satellite
images~\cite{martinis2010automatic,mason2012near,mason2010flood}. On
the one hand, algorithms based on earth still cameras cover a limited
area but do not have hardware and software restrictions. On the other
hand, algorithms based on satellite imagery cover a much wider area,
but if the system is to be installed on board the satellite, there are
many factors that have to be considered when deciding the type of
hardware and software to be used on board a satellite \cite{kogan1995droughts,mason2012near,song2013drought}.

\minisection{Drought:} droughts are significant weather-related
disasters that can result in devastating agricultural, health,
economic and social consequences. There are two main characteristics that determine the presence of drought in an area: a low surface soil moisture and a decrease in rainfall. Soil moisture data is rarely computed with visual data, however in \cite{hassan2015assessment} they use high-resolution remotely sensed data to determine the water balance of the soil. Another technique used to determine the beginning of drought is to compute the difference between the average precipitation, or some other climatic variable, over some time period and comparing the current situation to a historical record. Observation of rainfall
levels can help prevent or mitigate drought. Although rainfall does not
have a clear visual consequence that can be monitored through cameras,
there have been some image processing algorithms studied to map annual
rainfall records~\cite{fu2012drought,song2013drought}. These
algorithms help to determine the extension and severity of a drought
and to determine regions likely to suffer from this emergency.

\minisection{Earthquake:} earthquakes are sudden shakes or vibrations of the ground caused by tectonic movement. The magnitude of an earthquake can vary greatly from almost imperceptible earthquakes (of which hundreds are registered every day) up to very violent earthquakes that can be devastating. Earthquake prediction is a whole field of research called seismology. However, due to the lack of distinctive visual information about an earthquake, this emergency has received relatively little attention from the computer vision community. Several works address the robustness of motion tracking techniques under seismic-induced motions~\cite{doerr2005methodology,hutchinson2004monitoring},  with the aim of assisting rescue and reconnaissance crews. The main difficulty for these systems is the fact that the camera itself is also undergoing motion. In~\cite{doerr2005methodology} this problem is addressed by introducing a dynamic camera reference system which is shared between cameras and allows  tracking of objects with respect to this reference coordinate system. Also, computer vision plays a role during the rescue of people affected by these natural disaster by detecting people with robots or UAVs in the damaged areas, this is tackled in the ``injured civilians'' section.

\minisection{Hurricane/tornado:} hurricanes and tornadoes are violent storms characterized by heavy wind and rain. Similarly to earthquakes, depending on their magnitude they can be almost unnoticeable or devastating and at the same time the visual information derived from these natural disasters has not yet been studied in the field of computer vision. However, similarly to earthquakes, robots and UAVs can be used to find people in the areas affected by the disaster using human detection algorithms.

The more general work on ``injured civilians'' detection in affected areas which is shared among any other emergency that involves people rescue like earthquakes and hurricanes, is again applicable to landslide and avalanche emergencies. We found one work which evaluates this in particular for avalanches, and which takes into account the particular environment where avalanches occur~\cite{bejiga2017convolutional}. 



\subsection{Emergencies caused by humans}
Emergencies caused by humans cover adverse events and disasters that
result from human activity or man-made hazards. Although natural
emergencies are considered to generate a greater number of deaths and
have larger economic impact, emergencies caused by humans are
increasing in importance and magnitude. These emergencies can be
unintentionally caused, such as road accidents or drowning persons,
but they can also be intentionally provoked, such as armed
conflicts. In this survey, emergencies caused by humans are
divided in two types depending on their scope: emergencies affecting
multiple persons and emergencies affecting a single person.

\subsubsection{Emergencies affecting multiple persons}
These emergencies are human-caused and affects several persons.

\minisection{Road accident emergencies:} Although depending on the
extent of the emergency they can affect one or multiple persons, we
consider road accidents to be multiple-person emergencies. Transport
emergencies result from human-made vehicles. They occur when a vehicle
collides with another vehicle, a pedestrian, animal or another road
object. In recent years, the number of vehicles on the road has
increased dramatically and accidents have become one of the highest
worldwide death factors, reaching the number one cause of death among those aged between 15 and 29 \cite{world2015global}. In order to prevent these
incidents, safety infrastructure has been developed, many monitoring
cameras have been installed on roads, and different sensors have been
installed in vehicles to prevent dangerous situations. In computer
vision there are three main fields of study in which have focused on
the detection or prevention of transport emergencies.
\begin{itemize}
\item Algorithms which focus on detecting anomalies, as in most cases
  they correspond to dangerous
  situations~\cite{jiang2011anomalous,saligrama2012video,sultani2010abnormal}.
\item Algorithms specialized on detecting a concrete type of road
  emergency, such as a collision~\cite{kamijo2000traffic,veeraraghavan2003computer,yun2014motion}.
\item Algorithms based on on-board cameras which are designed to
  prevent dangerous situations, such as algorithms that detect
  pedestrians and warn the
  driver~\cite{gavrila2007multi,guo2012robust,hachisuka2011facial}.
\end{itemize}

\minisection{Crowd-related emergencies:} These type of emergencies are
normally initiated by another emergency when man persons are gathered
in a limited area. They normally occur in multitudinous events such as
sporting events, concerts or strikes. These type of events pose a
great risk because when any emergency strikes it can lead to
collective panic, thus aggravating the initial emergency. This is a
fairly new area of research in computer vision. In computer vision for
crowd-related emergencies we find different sub-topics like crowd
density estimation, analysis of abnormal
events~\cite{andrade2006detection}, crowd motion
tracking~\cite{ihaddadene2008real} and crowd behaviour
analysis~\cite{garate2009crowd}, among others.

\minisection{Weapon threat emergencies:} Although detecting weapons
through their visual features and characteristics has not yet been
studied in computer vision, some studies have been done in the
direction of detecting the heat produced by them. A dangerous weapon
that threatens innocent people every year are mines. Millions of old
mines are hidden under the ground of many countries which kill and
injure people. These mines are not visible from the ground but they
emit heat that can be detected using thermal
cameras~\cite{muscio2004land,siegel2002land,wasaki2001smart}. Another
weapon that can be detected through thermal sensors are guns, due to
the heat that they emit when they are fired~\cite{price2004system}.

\subsubsection{Emergencies affecting a single person}
These are emergencies are caused by or affect one person. The classic
example of this is detecting human falling events for monitoring
the well-being of the elderly.

\minisection{Drowning:} Drowning happens when a person lacks air due
to an obstruction of the respiratory track. These situations normally
occur in swimming pools or beach areas. Although most public swimming
pools and beaches are staffed with professional lifeguards, drowning
or near-drowning incidents still occur due to staff reaction
time. Therefore there is a need to automate the detection of these
events. One way of automating detection of these events is the
installation of monitoring cameras equipped with computer vision
algorithms to detect drowning persons in swimming pools and beach
areas. However, drowning events in the sea are difficult to monitor
due to the extent of the hazardous area, so most computer vision
research on this topic has been focused in drowning people in swimming
pools~\cite{eng2003automatic,zecha2012swimmer}.

\minisection{Injured person:} An injured person is a person that has
come to some physical harm. After an accident, reaction time is
crucial to minimize human losses. Emergency responders focus their
work on the rescue of trapped or injured persons in the shortest time
possible. A method that would help speed up the process is the use of
aerial and ground robots with visual sensors that record the affected
area at the same time as detect human victims. Also, in the event
of an emergency situation, knowing if there are human victims in the
area affected by the event may help determine the risk the responders
are willing to take. For this purpose autonomous robots have been
built to automatically explore the area and detect human victims
without putting at risk the rescue
staff~\cite{andriluka2010vision,castillo2005method,kleiner2007genetic,leira2015automatic,rudol2008human,soni2013victim}.
Detecting injured or trapped humans in a disaster area is an extremely
challenging problem from the computer vision perspective due to the
articulated nature of the human body and the uncontrolled environment.

\minisection{Falling person:} With the growth of the elderly
population there is a growing need to create timely and reliable smart
systems to help and assist seniors living alone. Falls are one of the
major risks the elderly living alone, and early detection of falls is
very important. The most common solution used today are call buttons,
which when the senior can press to directly call for help. However
these devices are only useful if the senior is conscious after the
fall. This problem has been tackled from many different perspectives.
On the one hand wearable devices based on accelerometers have been
studied and developed~\cite{chen2005wearable,zhang2006fall}, some of
these devices are in process of being commercialized
\cite{fallDet2015}. On the other hand, falling detection algorithms
have been studied from a computer vision
perspective~\cite{liao2012slip,rougier2011fall}.

\section{Monitoring Objective} 
\label{sec:objective}

As stated in Section \ref{sec:organization}, an emergency has four main phases which at the same time lead to four emergency management phases: prevention, detection, response/assistance and understanding of emergencies. It is important to understand, study and identify the different phases of an emergency life cycle since the way to react to it changes consequently. In this section, we will classify computer vision algorithms for emergency situations depending on the emergency management phase in which they are used.

\begin{figure*}[t!]
\begin{center}
\begin{tabular}{ccc}
\subfloat[Example of fire prevention by detecting smoke. Example taken from \cite{gubbi2009smoke}.]{\includegraphics[width = 0.3\linewidth]{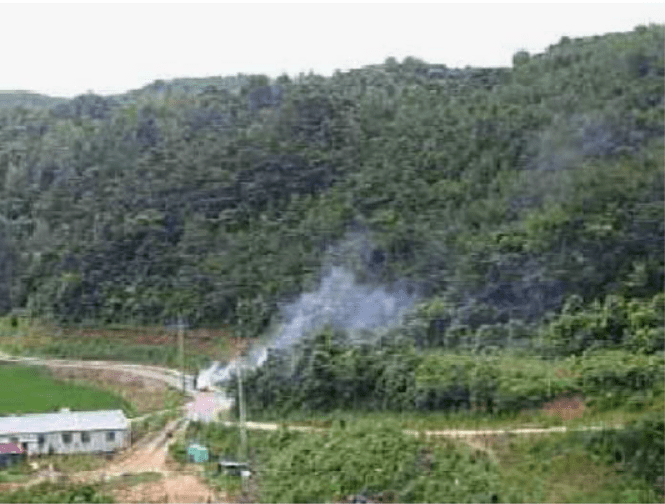}}&
\subfloat[Example of pedestrian detection using a vehicle on-board camera to prevent pedestrians being run over \cite{wojek2009multi}.]{\includegraphics[width = 0.3\linewidth]{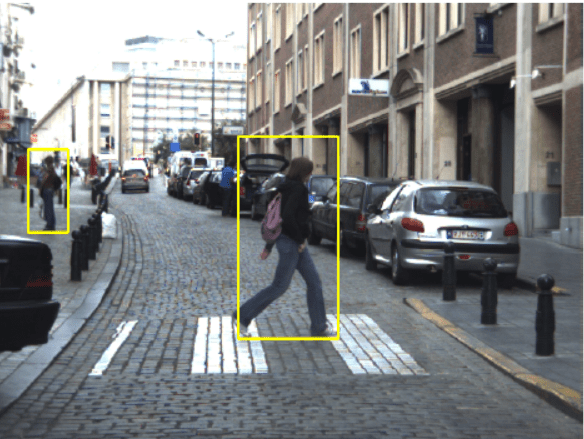}} &
\subfloat[Example of drowsy driver detection taken from \cite{choi2014head}.]{\includegraphics[width = 0.3\linewidth]{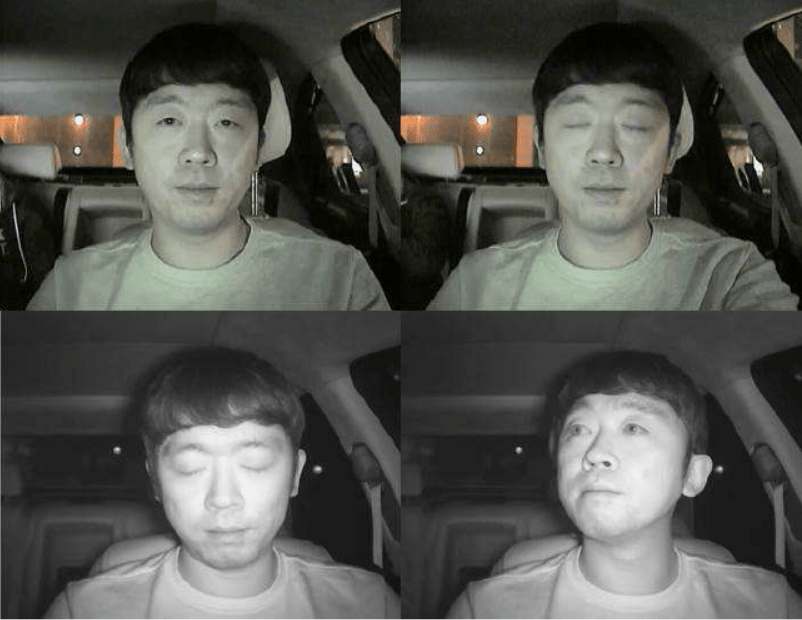}}
\end{tabular}
\caption{Example situations in which computer vision techniques have been applied for emergency prevention.}
\label{fig:example prevention}
\end{center}
\end{figure*}

\subsection{Emergency Prevention}
\label{subsec:prevention}

This phase focuses on preventing hazards and risk factors to minimize the likelihood that an emergency will occur. The study of emergency prevention on computer vision can not be applied to any emergency since not all of them are induced by risk factors that have relevant visual features that can be captured by optical sensors. For example, the only visual feature which is normally present before a fire starts is smoke. Therefore research on fire prevention in computer vision focuses on smoke detection. As an example, in Figure \ref{fig:example prevention} (a) we see a region of forest with a dense smoke that looks like the starting of a fire, however a fire detection algorithm would not detect anything until the fire is present in the video. Some early work on this topic started at the beginning of the century \cite{chen2004early,gomez2003smoke}. Since then, smoke detection for fire prevention or early fire detection has been an open topic in computer vision. Most of the work done on smoke detection is treated as a separate problem from fire detection and aims to detect smoke at the smoldering phase to prevent fire \cite{gubbi2009smoke,xiong2007video,ye2015dynamic,yuan2011video}. However there is also some work on smoke and flame detection when the fire has already started since smoke in many cases can be visible from a further distance \cite{kolesov2010natural,yu2013real}.

Traffic accidents are considered as one of the main sources of death and injuries globally. Traffic accidents can be caused by three main risk factors: the environment, the vehicle and the driver. Among these risk factors, it is estimated that the driver accounts for around 90\% of road accidents \cite{Traffic2015}. To avoid these type of accidents, a lot of security devices have been incorporated on vehicles in the past years. There are two types of security devices: ``active'' whose function is to decrease the risk of an accident occurring, and ``passive'' which try to minimize the injuries and deaths in an accident. Among the active security devices, there is work on the usage of visual sensors to detect situations that could lead to an accident and force actions that could be taken to prevent the accident \cite{trivedi2007looking}.

There are many different kind of traffic accidents but one of the most common and at the same time more deadly ones are pedestrians being run over. Due to the variant and deformable nature of humans, the environment variability, the reliability, speed and robustness that a security system is required, the detection of unavoidable impact with pedestrians has become a challenge in computer vision.  This field of study is well defined and with established benchmarks and it has served as a common topic in numerous review articles \cite{benenson2014ten,dollar2012pedestrian,geronimo2009survey,wojek2009multi}.  Now, with the increase of training data and computing power this investigation has had a huge peak in the past years and even the first devices using this technology have already been commercialized \cite{2014mobileeye,2016flir}. Figure \ref{fig:example prevention} (b) shows an example of the output from a pedestrian detection algorithm on board a vehicle.

Another contributing factor to many accidents is the drivers fatigue, distraction or drowsiness. A good solution to detect the state of the driver in a non intrusive way is using an on board camera able to detect the drivers level of attention to the road. There are different factors that can be taken into account to determine the level of attention to the road of a driver. In \cite{choi2014head} they determine if the driver is looking at the road through the head pose and gaze direction, in Figure \ref{fig:example prevention} (c) we show some examples of the images they used to train their algorithm. While in \cite{hachisuka2011facial} they determine 17 feature points from the face that they consider relevant to detect the level of drowsiness of a person. In \cite{hu2009driver} they study the movements of the eyelid to detect sleepy drivers.

Although not so widely studied, there are other type of road accidents that could also be prevented with the help of computer vision. For example, in \cite{hayashi2009predicting} they propose a framework to predict unusual driving behavior which they apply to right turns at intersections. Algorithms that use on board cameras to detect the road have also been studied \cite{guo2012robust}, this could help to detect if a vehicle is about to slip on the road. In \cite{barth2010tracking} they propose a system with on board stereo cameras that assist drivers to prevent collisions by detecting sudden accelerations and self-occlusions. Multiple algorithms for traffic sign detection with on board cameras have also widely proposed in the literature \cite{houben2013detection,mogelmose2014traffic}, these algorithms mostly go in the direction of creating autonomous vehicles, however they can also be used to detect if the driver is not following some traffic rules and is therefore increasing the risk of provoking an accident.

In order to prevent falls, in \cite{liao2012slip} they propose a method to detect slips arguing that a slip is likely to lead to a fall or that a person that just had a slip is likely to have balance issues that increase the risk of falling down in the near future. They consider a slip as a sequence of unbalanced postures.


\begin{figure*}[t!]
\begin{center}
\begin{tabular}{cc}
\subfloat[Two very different scenarios such as sitting and falling may be visually similar, in order to differentiate them some researchers suggest to use temporal information \cite{chua2015simple}.]{\includegraphics[height = 4.5cm]{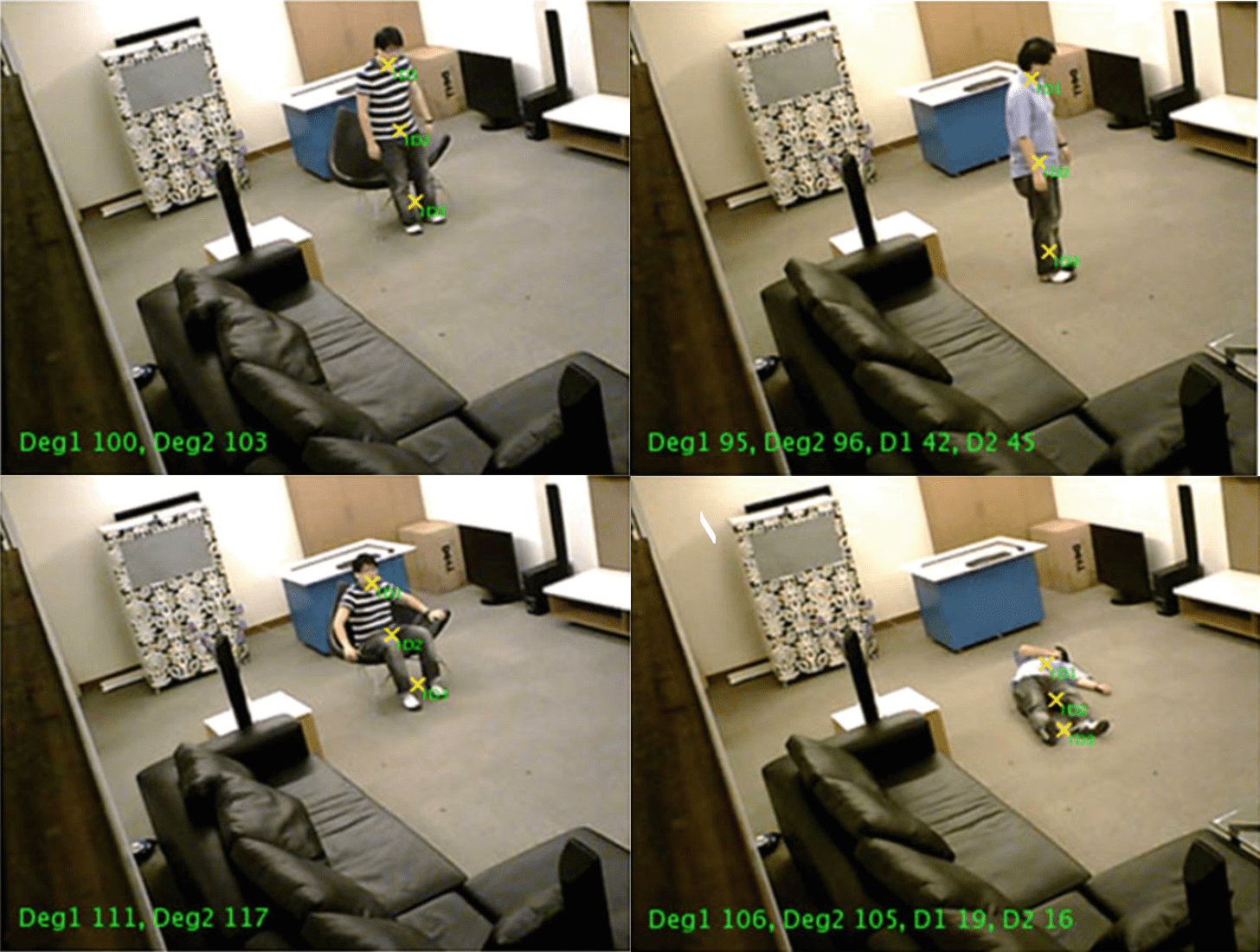}}&
\subfloat[Example of a person detection algorithm in a swimming pool  \cite{eng2003automatic}.]{\includegraphics[height = 4.5cm]{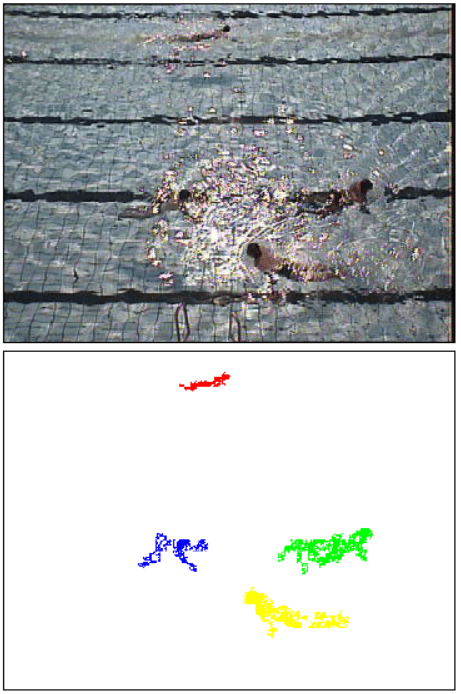}}
\end{tabular}
\caption{Example situations in which computer vision techniques have been applied for emergency detection.}
\label{fig:example detection}
\end{center}
\end{figure*}

\subsection{Emergency Detection}

The field of emergency detection through computer vision in contrast with the field of prevention is much wider and has been studied in more depth. In this section we will present the visual information that is being studied to detect different emergencies.

Fire detection has been widely studied in computer vision since it is very useful on open spaces, where other fire detection algorithms fail. In some cases it is useful to not only detect the fire but also some properties of the fire such as the flame height, angle or width \cite{martinez2008computer}. In order to detect fire on visual content, the algorithms studied exploit visual and temporal features of fire. In early research, flame was the main visual feature that was studied in fire detection \cite{celik2009fire,gunay2010fire,ko2010early,rinsurongkawong2012fire,truong2012fire,van2010fire} however, as stated in Section \ref{subsec:prevention}, recently smoke has also been studied in the field of fire detection mainly because it appears before fire and it spreads faster so in most cases smoke will appear faster in the camera field of view \cite{kolesov2010natural,yu2013real}. As fire has a non-rigid shape with no well defined edges, the visual content used to detect fire is related to the characteristic color of fire. This makes the algorithms in general vulnerable to lighting conditions and camera quality.

Water, the main cue for flood detection, has similar visual difficulties as fire in order to be detected since it has a non-rigid shape and the edges are also not well defined. Moreover, water does not have a specific characteristic color because it might be dirty and it also reflects the surrounding colors. Due to the mentioned difficulties flood detection in computer vision has not been so widely studied. In \cite{lai2007real} they use color information and background detail (because as said before the background will have a strong effect on the color of the water) and the ripple pattern of water. In \cite{borges2008probabilistic} they analyse the rippling effect of water and they use the relationship between the averages for each color channel as a detection metric.

Detecting person falls has become a major interest in computer vision, specially motivated to develop systems for elderly care monitoring at home. In order to detect a falling person it is important to first detect the persons which are visible in the data and then analyse the movements. Most algorithms use background subtraction to detect the person since in indoor situations, most of the movement come from persons. Then, to determine if the movement of the person corresponds to a fall some algorithms use silhouette or shape variation \cite{liao2012slip,mirmahboub2013automatic,rougier2007fall}, direction of the main movement \cite{hazelhoff2008video}, or relative height with respect to the ground \cite{mogelmose2014traffic}. However, sometimes it is difficult to distinguish between a person falling down and a person sitting, lying or sleeping, to address this issue, most systems consider the duration of the event since a fall is usually a faster movement \cite{lin2007automatic,mogelmose2014traffic}, for example, in \cite{chua2015simple} they consider 10 frames as a fast movement and use this threshold to differentiate two visually similar movements such as the ones given in Figure \ref{fig:example detection} (a) which corresponds to sitting and falling.

The problem of drowning person detection is fairly similar to the falling person detection, since it is also in a closed environment and related to human motion. Similarly to fall detection algorithms, the first step to detect a drowning person is to detect the persons visible in the camera range, which is also done by background subtraction \cite{fei2009drowning,kam2002video,lu2004vision,zecha2012swimmer}. In Figure \ref{fig:example detection} we give an example of a detection of the persons in a swimming pool using a background modelling algorithm. However, in this case aquatic environments have some added difficulties because it is not static \cite{chen2011framework}.

\begin{table*}[t!]
\footnotesize \baselineskip 0.5 pt
\begin{tabular}{l*{4}{c}l}
\toprule
& \multicolumn{4}{c}{\bf Sensor} \\
\cmidrule(lr){2-5}
{\bf Location}& \parbox{3 cm}{\bf Monocular  camera} & \parbox{3cm}{\bf Infrared/ Thermal \\ camera} & \parbox{3cm}{ \bf Stereo  camera} & \parbox{3 cm}{\bf Other}\\
\addlinespace[.3 em]   \midrule
  Fixed location & \parbox{3cm}{- \$50 to \$2000 \\
- Up to 30 MP\\
- Real time data\\ acquisition\\
- Limited area\\ of visualization\\
}&  \parbox{3cm}{\vspace{0.1cm} - \$ 200 to \$ 6000\\ - Up to 1.2 MP\\- Up to 0.02\degree C\\ sensitivity\\ - Real time data\\ acquisition \\ - Limited area of\\ visualization\\ } & \parbox{2.3cm}{- \$ 200 to \$ 2000\\ - Up to 30 MP\\ - Real time data\\ acquisition \\ - Limited area of\\ visualization\\  } &\\
\midrule
Robot or UAV  & \parbox{3cm}{- \$100 to \$2000 \\
- Up to 30 MP \\
- Portable \\
- Bad quality\\data acquisition\\ in real time \\
- Limited \\autonomy\\}&  \parbox{3cm}{\vspace{0.2cm} - \$250 to \$2000\\
- Up to 1.2 MP\\
- Up to 0.02\degree C\\ sensitivity\\
- Portable \\
- Bad quality \\data acquitision \\in real time\\} & \parbox{3cm}{- \$ 250 to \$ 2000 \\
- Up to 30 MP \\
- Portable \\
- Bad quality \\data acquitision \\in real time \\
- Limited auton-\\omy} &\\
\midrule
Satellite & & & & \parbox{3 cm}{\vspace{0.1cm}- Open data access through many research programs \\
- Spatial \\resolution of around 10m \\
- Can cover the entire globe \\
- May have to wait a few days until the satellite covers the entire globe\\
- Image acquisition is influenced by weather conditions like clouds}\\
\addlinespace[.3 em]  \bottomrule
\end{tabular}
\caption{Summary of the advantages and disadvantages of the types of sensors and the location at which they can be placed.}
\label{tab:sensors}
\end{table*}

\subsection{Emergency Response/Assistance}

After an emergency has already occurred, one of the most important tasks of emergency responders is to help injured people. However, depending on the magnitude of the disaster, it may be difficult to effectively localize all these persons so emergency responders may find it useful to use robots or UAVs to search for these persons. In order to automatically detect injured people, some of these robots use cameras together with computer vision algorithms. When considering disasters taking place on the ground one of the main characteristic of an injured person is that they are normally lying on the floor so these algorithms are trained to detect lying people. The main difference between them is the perspective from which these algorithms expect to get the input, depending if the camera is mounted on a terrestrial robot \cite{castillo2005method,kleiner2007genetic,soni2012classifier,soni2013victim} or on a UAV \cite{andriluka2010vision,leira2015automatic,rudol2008human}.

A fast and efficient response the in the moments after an emergency has occur ed is critical to minimize damage. For that it is essential that emergency responders have situational awareness, which can be obtained through sensors in the damaged areas, by crowd sourcing applications, and from social media. However, extracting useful information from huge amounts of data can be a slow and tedious work. Therefore, organization of video data that facilitates the browsing on the data is very important \cite{schoeffmann2015video}. Since during an emergency it is common to have data on the same event from different angles and perspectives, in \cite{tompkin2012videoscapes} the authors combine interactive spatio-temporal exploration of videos and 3D reconstruction of scenes. In \cite{huang2014videoweb} the authors propose a way of representing the relationship among clips for easy video browsing. To extract relevant information from videos, in \cite{diem2016video} the authors find recurrent regions in clips which is normally a sign of relevance. In \cite{lopez2017bandwidth} the authors introduce a framework of consecutive object detection from general to fine that automatically detects important features during an emergency. This also minimizes  bandwidth usage, which is essential in many emergency situations where critical communications infrastructure collapses.

\subsection{Emergency Understanding}

Keeping record of emergencies is very important to analyse them and extract information in order to understand the factors that have contributed to provoke the emergency, how these factors could have been prevented, how the emergency could have been better assisted to reduce the damages or the effects of the emergency. The systems developed to with this objective should feed back into the rest of the systems in order to improve them.

\section{Acquisition} 
\label{sec:acquisition}

Choosing an appropriate acquisition setup is one of the important design choices. To take this decision one must take into account the type of emergency addressed (see Section~\ref{sec:Definition}) and objectives to fulfill (see Section~\ref{sec:objective}). One of the things to consider is the location where the sensor will be placed (e.g. in a fixed location, mounted on board a satellite, a robot, or a UAV, etc). Then, depending on the type of sensor we acquire different types of visual information: monocular, stereo, infrared, or thermal. Depending on the frequency in which the visual information is acquired we can obtain videos or still images. Finally, this section is summarized in Table \ref{tab:sensors} where we include the advantages and disadvantages of the types of sensors studied and the locations at which they can be placed The type of sensors chosen for the system will have an important impact on the types of algorithms that can be applied, the type of information that can be extracted and the cost of the system, which is also many times a limitation.



\subsection{Sensor location}
The physical location where the sensor is installed determines the
physical extent in the real world that will be captured by the
sensor. The location of the sensor depends primarily on the expected
physical extent and location of the emergency that is to be prevented,
detected, assisted or understood.

\minisection{In a fixed location:}
\label{minisec:static} When the type of emergency considered has a
limited extent and the location where it can occur (the area of risk)
is predictable, a fixed sensor is normally installed. In many cases
this corresponds to a monitoring camera mounted on a wall or some
other environmental structure. These are the cheapest and most
commonly used sensors. They are normally mounted in a high location,
where no objects obstruct their view and pointing to the area of
risk.

Static cameras have been used to detect drowning persons in swimming
pools, as in this case the area of risk is limited to the swimming
pool
\cite{chen2010hidden,chen2011framework,eng2003automatic,fei2009drowning,kam2002video,lu2004vision,zecha2012swimmer}. They have also been used to monitor dangerous areas on roads, inside
vehicles either pointing to the driver to estimate drowsiness
\cite{choi2014head,hachisuka2011facial,kumar2009application,lee2011real},
or pointing to the road to process the area in which it is being driven
\cite{geronimo2009survey,guo2012robust,hayashi2009predicting,houben2013detection,ma2015real,mogelmose2014traffic}. Fixed
cameras are also widely used in closed areas to detect falling persons \cite{chua2015simple,jiang2013real,mirmahboub2013automatic,rougier2007fall}. Finally, they can occasionally also be used to
detect floods or fire in hotspots \cite{borges2008probabilistic,celik2009fire,lai2007real,rinsurongkawong2012fire}.

\minisection{On board robots or UAVs:} For emergencies that do not
take place in a specific area, or that need a more immediate
attention, sensors can be mounted on mobile devices such as robots or
UAVs. The main difference between these two types of devices is the
perspective from which the data can be acquired: terrestrial in the
case of a robot, and aerial for UAVs. On the one hand, the main
advantage of these data sources is that they are flexible and support
real time interaction with emergency responders. On the other hand,
their main drawback is their limited autonomy due to battery
capacity. Robots and UAVs have been used extensively for human
rescue. In such cases, a visual sensor is mounted on board the
autonomous machine which can automatically detect human
bodies~\cite{andriluka2010vision,castillo2005method,leira2015automatic,soni2012classifier,soni2013victim}.
The authors of~\cite{hassan2015assessment} use a UAV with several
optical sensors to capture aerial images of the ground, estimate the
soil moisture, and determine if an area suffers or is close to
suffering drought.

\minisection{On board satellites:} For emergencies of a greater
extent, such as drought, floods or fire, sometimes the use of sensors
on board satellites is more convenient \cite{hassan2015assessment,mason2012near,mason2010flood,song2013drought}. By using satellites, it is
possible to visualize the whole globe. However, the biggest drawback
of this type of sensor is the frequency at which the data can be
acquired since the satellites are in continuous movement and it may be
necessary to wait up to a few days to obtain information about a
specific area, as an example the Sentinel 2 which is a satellite in the Copernicus program for disaster management revisits every location within the same view angle every 5 days \cite{sentinel2esa}. Due to this, the information acquired by this type of
sensor is mainly used for assistance or understanding of an emergency
rather than to prevent or detect it.

\minisection{Multi-view cameras:} In some scenarios, like in the street, it is useful to have views from different perspectives in order to increase the range or performance of the algorithm. However, using data from different cameras also entails some difficulties in combining the different views. Since the most common scenario where multiple cameras are present is in the street, this problem has been tackled in pedestrian detection \cite{utasi2012multi} and traffic accidents \cite{forczmanski2016multi}. 

\subsection{Type of sensor}

Another important decision to take when working with computer vision
for emergency management is the type of optical sensor that should be
used. When taking this decision it is important to take into account
the technical characteristics of the available sensors, the type of
data that they will capture, and the cost.

\minisection{Monocular RGB or greyscale cameras:} Monocular cameras
are devices that transform light from the visible spectrum into an
image. This type of visual sensor is the cheapest and therefore by far
the most widely used in computer vision and monitoring
systems. Although these sensors may not be the optimal in some
scenarios (such as fire detection), due to their low cost and the
experience of researchers on the type of images generated by such
sensors, they are used in almost all emergencies \cite{anderson2009falling,eng2003automatic,hu2006system,lai2007real,mehran2009abnormal,toreyin2006computer}.

\minisection{Stereo RGB or greyscale cameras:} Stereo cameras are
optical instruments that, similarly to monocular cameras, transform
light from the visible spectrum into an image. The difference between
stereo and monocular cameras is that stereo sensors have at least two
lenses with separate image sensors. The combination of the output from
the different lenses can be used to generate stereo images from which
depth information can be extracted. This type of sensor are especially
useful on board vehicles where the distance between the objects and the
vehicle is essential to estimate if a collision is likely to occur and
when~\cite{barth2010tracking}. They are also used on board of vehicles
to detect lanes~\cite{danescu2009probabilistic,guo2012robust}.

\minisection{Infrared/thermal cameras:} Infrared or thermal cameras
are devices that detect the infrared radiation emitted by all objects
with a temperature above absolute
zero~\cite{gade2014thermal}. Similarly to common cameras that
transform visible light into images, thermal cameras convert infrared
energy into an electronic signal which is then processed to generate a
thermal image or video. This type of camera has the advantage that
they eliminate the illumination problems that normal RGB or greyscale
cameras have. They are of particular interest for emergency management
scenarios such as fire detection since fire normally has a temperature
higher than its environment.

It is also well known that humans have a body temperature that ranges
between 36$^{\circ}$C and 37$^{\circ}$C in normal conditions. This
fact can be exploited to detect humans, and has been used for
pedestrian detection to prevent vehicle collisions with
pedestrians~\cite{suard2006pedestrian,wang2012pedestrian}, to detect
faces and pupils~\cite{kumar2009application}, and for fall detection
for elderly monitoring~\cite{sixsmith2004smart}. Although not so
extensively used, thermal and infrared cameras have also been used for
military purposes, such as to detect firing guns
\cite{price2004system} and to detect mines under ground
\cite{muscio2004land,siegel2002land,wasaki2001smart}.

One of the disadvantages of thermal cameras compared with other type
of optical sensors is that their resolution is lower than many other
visual sensors. For this reason some systems use thermal cameras in
combination with other sensors. The authors
of~\cite{hassan2015assessment} use a thermal camera among other visual
sensors to study the surface soil moisture and the water balance of
the ground, which is in turn used to estimate, predict and monitor
drought.

In order to obtain depth information in addition to thermal
information, it is also possible to use infrared stereo cameras such
as in~\cite{bertozzi2005infrared} where the authors use infrared
cameras to detect pedestrians by detecting their corporal heat, and
then estimate pedestrian position using stereo information.

\subsection{Acquisition rate}

Depending on the rate at which optical data is acquired we can either
obtain video or still images. The main difference between video and a
still image is that still images are taken individually while video is
an ordered sequence of images taken at a specific and known rate. To
be considered a video, the rate at which images are taken must be
enough for the human eye to perceive movement as continuous.

\minisection{Video:} The main advantage of videos over images is that,
thanks to the known acquisition rate, they provide temporal
information which is invaluable for some tasks. Also, most of the
systems studied in this survey are monitoring systems meant to
automatically detect risky situations or emergencies. As these
situations are rare but could happen at any moment and it is important
to detect them as soon as possible, so it is necessary to have a
high acquisition rate.

For the study of many of the emergencies considered in this survey it
is necessary to have temporal information. For example, merely
detecting a person lying on the ground in a single image is not
sufficient to infer that that person has \emph{fallen}.  Many systems
distinguish between a person lying on the floor and a person that has
fallen by using information about the speed at which this person
reached the final
posture~\cite{lin2007automatic,rougier2011fall}. Also, systems that
estimate likelihood of a car crash must have temporal information to
know at which speed the external objects are approaching the
vehicle. To estimate driver drowsiness or fatigue, many systems take
information into account such as the frequency at which the driver
closes his eyes and for how long they remain closed
\cite{kumar2009application}.

Moreover, many algorithms developed for the detection of specific
situations also require temporal information. In order to detect fire,
most algorithms exploit the movement of the fire to differentiate it
from the background. Similarly, for drowning person detection, many
algorithms use temporal information to easily segment the person from
the
water~\cite{chen2011framework,eng2003automatic,fei2009drowning,lu2004vision}. Also,
many falling person detection systems use tracking algorithms to track
the persons in the scene~\cite{hazelhoff2008video} or background
subtraction algorithms to segment persons~\cite{mirmahboub2013automatic}.

\minisection{Still images:} Still images, in contrast with video, do
not give information about the movement of the objects in the scene
and does not provide any temporal information. Although most of the algorithms that study the problem of fire detection use temporal information and therefore require video, David Van et al. \cite{van2010fire} study  a system that detects fire on still images to broaden the applicability of their system. In some cases due to the sensor used by the system it is not possible or convenient to obtain videos for example in systems that use satellite imagery \cite{hassan2015assessment,mason2012near,mason2010flood,song2013drought}, also in systems that use UAVs where in many cases is more convenient to take still images instead of videos in order to effectively send the data in real time or get higher resolution imagery \cite{castillo2005method,soni2012classifier}.

\subsection{Sensor cost}
When implementing the final system, another important factor that should be taken into account before designing it is the cost from setting up the system. In general, the cost of the final system is going to be the cost of the visual sensor, the cost of the installation of the system and the cost of the data transmission. In this section we will do a brief study of the three sources of costs.

\begin{table*}[t]
\begin{center}
\begin{tabular}{p{3cm}p{8cm}}
\toprule
\hspace{1cm} {\bf Device} & {\bf Description} \\
\toprule
\raisebox{-1.5cm}{\includegraphics[width=3cm]{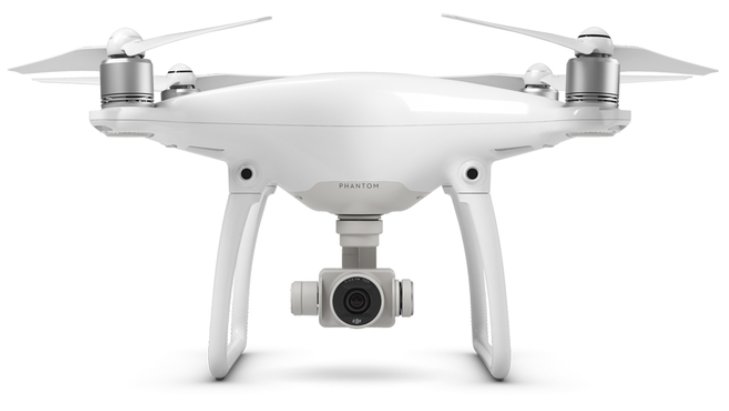}}& Example of a commonly used UAV. This UAV has an integrated camera of 12 Mpx, a flying autonomy of 28 minutes, a maximum transmission distance of 5km a live view working frequency of 2.4GHz and a live view quality of 720P and 30fps. This device can be bought on the market for around \$1200.\\
\midrule
\hspace{0.45cm} \raisebox{-2.75cm}{\includegraphics[width=2cm]{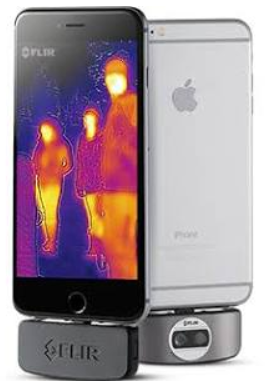}} & \vspace{0.5cm} Example of a thermal camera that can be attached to a smart phone. It has an VGA resolution a sensibility of 0.1ºC and works with temperatures between -20ºC to 120ºC. A device if these characteristics can be bought for around \$300.\\
\midrule
\hspace{0.25cm}\raisebox{-1.25cm}{\includegraphics[width=2.5cm]{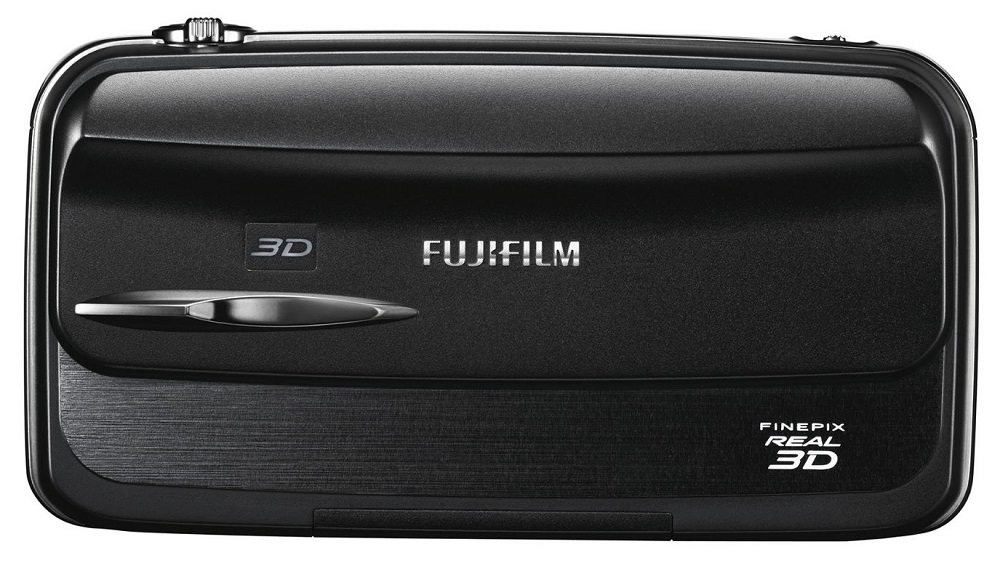} }& \vspace{0.25cm} Example of a built in stereo camera of 10 MP resolution and a cost of \$600.\\
\bottomrule
\end{tabular}
\end{center}
\caption{Some examples of devices used in computer vision for
  emergency management systems, their cost and specifications.}
\label{tab:table of types of sensors}
\end{table*}

The most common visual sensors are monocular RGB and grayscale cameras. Since these sensors are so common and have a big demand, their technical features have improved greatly over time and their price has dramatically decreased. Therefore there exists a wide range of monocular cameras in the market with different technical features and prices. Depending on the quality of the device, its price can vary between approximately \$50 to \$2000. Although there also exists a wide variety of infrared cameras, these devices are less common and in general more expensive. Depending on the quality of the camera, its price fluctuates between \$200 and \$6000. To illustrate that, in the second row from Table \ref{tab:table of types of sensors} we give an example of a thermal camera of \$300 and its specifications. Finally, stereo cameras are the less common visual sensors and there is fewer variety of these devices. Their price oscillates between \$200 and \$2000. In the third row from Table \ref{tab:table of types of sensors} we present an example of an average stereo camera. However, it is fairly common to build a stereo device by attaching two monocular cameras which makes its price double the price of a monocular camera.

The cost of the installation of the system and the transmission of data varies greatly depending on the localization in which the sensor is installed. In this case, the cheapest installation would be at a fixed location, for which it would only be necessary to buy a holder to fix the sensor and acquire a broadband service in order to transmit the data in real time. On the other hand, it is also possible to install the sensor on a mobile device such as a robot or UAV. In this case it is necessary to first buy the device, whose price also have big fluctuations depending on their quality and technical characteristics such as its autonomy, dimensions and stability. A low cost robot or UAV can be bought for around \$50 while high quality ones can cost around \$20000. Moreover, these devices require a complex transmission system since the sensor is going to be in continuous movement. In fact, since these systems are meant to acquire data over a limited period of time, it is common not to install any transmission data system but saving all the data in the device and processing it after the acquisition is over. It is also possible to install a antenna on the device that transmits these data in real time over 2.4GHz or 5.8GHz frequency. In the first row from Table \ref{tab:table of types of sensors} we show an example of a commonly used drone with an integrated camera and with a 2.4GHz antenna for live data transmission.

\section{Algorithms}
\label{sec:algorithms}

\begin{figure*}
\centering
\includegraphics[width = 0.75\linewidth]{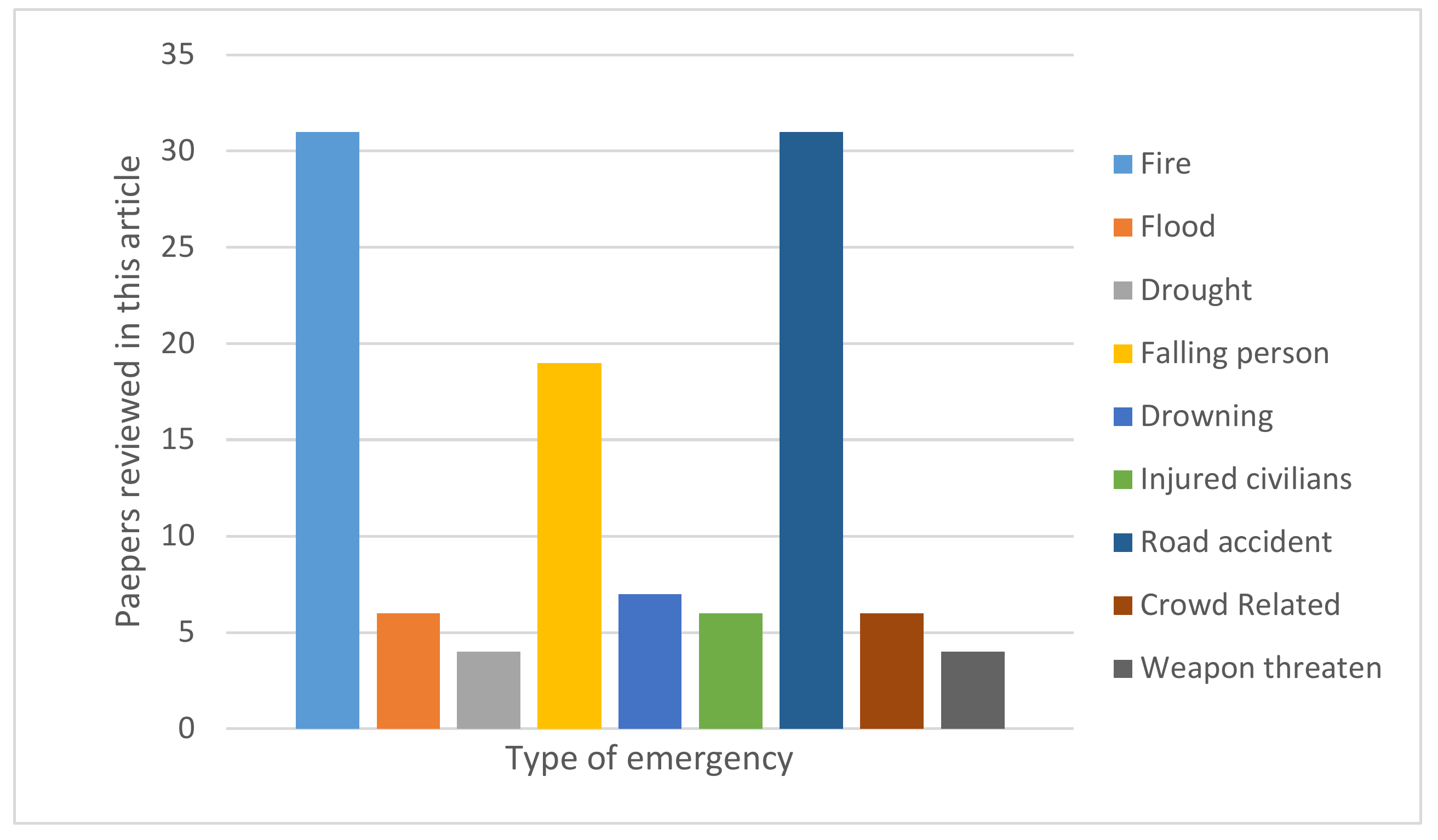}
\caption{Articles studied in this survey by type of emergency.}
\label{graph:types of emergencies}
\end{figure*}

In this section we review the most popular algorithms that have been
applied in computer vision applications for emergency situations. We have considered the nine different types of emergencies that we introduced in Section \ref{sec:Definition}. When selecting the articles to review in this survey we have considered a total of 114 papers that propose a computer vision system for some emergency management. However, not all the types of emergencies have bee equally studied in the literature nor in the survey, in Figure \ref{graph:types of emergencies} we show a bar graph of the distribution of the articles among the different types of emergencies.

This section is divided in two main parts: feature extraction and machine
learning algorithms. In the feature extraction part various approaches
to extracting relevant information from the data are elaborated. Once
that information is extracted, machine learning algorithms are applied
to incorporate spatial consistency (e.g. Markov Random Fields), to
model uncertainty (e.g. Fuzzy logic) and to classify the data to
high-level semantic information, such as the presence of fire or an
earthquake. We have focused on methods which are (or could be)
applied to a wider group of applications in emergency management.

\subsection{Feature extraction}
Feature extraction in computer vision is the process of reducing the
information in the data into a more compact representation which
contains the relevant information for the task at hand. This is a
common practice in computer vision algorithms related to emergency
situations. We divide the features in three main groups: color, shape
and temporal.

\subsubsection{Color features}
The understanding of emergency situations often involves the detection
of materials and events which have no specific spatial extent or
shape. Examples include water, fire, and smoke. However, these
textures often have a characteristic color, and therefore color
features can provide distinctive features.

Color analysis is a well established computer vision research field
and has been broadly applied in applications in emergency
management. It is frequently used to detect fire and smoke since they
have characteristic colors and -- although not so frequently -- it has
also been used to detect flooded regions. Fire color may vary
depending on the combustion material and the temperature. However most
flames are yellow, orange and especially red. Smoke tends to be dark
or white and can be more or less opaque. Water in flooded regions,
according to~\cite{borges2008probabilistic}, presents a smooth dark
surface or strong reflective spots much brighter than the
surroundings.


One of the principle choices in color feature extraction is the color
space or color model from which to compute the features. This choice
influences characteristics of the color feature, including photometric
invariance properties~\cite{gevers2012color}, device dependence, and
whether it is perceptually uniform or
not~\cite{ibraheem2012understanding}. Here we will shortly review the
most used color models and briefly describe some of their advantages
and disadvantages.




The data produced by camera devices is typically in \textbf{RGB} color
model. Due to the effortless availability of this color model and the
separation of the red values, this color model has been often used for
fire, smoke and flood detection
\cite{borges2010probabilistic,chen2010multi,rinsurongkawong2012fire,toreyin2006computer}. The
main disadvantage of this color model is its lack of illumination
invariance and dependence on the sensor. The \textbf{HSI} and \textbf{HSV} color models were proposed to resemble the
color sensing properties of human vision and separate the luminance
component from the chromatic ones.  They have been applied to smoke
detection~\cite{barmpoutis2014smoke}, to detect light gray
water~\cite{borges2008probabilistic} and to detect deep
water~\cite{lai2007real}. Several systems combine RGB and HSI color
models~\cite{chen2004early,rinsurongkawong2012fire,yu2013real}. The \textbf{YCbCr} color model has three components, one of which encodes the
luminance information and two encode the chromaticity. Having the
luminance component separated from chromatic components makes it
robust to changes in illumination \cite{celik2009fire,lai2007real}. Similarly to HSI and YCbCr, the \textbf{CIELab} color model makes it possible to separate the luminance information from the chromaticity.  Moreover, it is device independent and perceptually
uniform. This color model has been used to detect
fire~\cite{celik2010fast,truong2012fire} and to improve the background
models for swimming pools~\cite{eng2003automatic}.


\subsubsection{Shape and texture features}
There are elements which are very
characteristic of some of the emergencies presented in this survey,
such as smoke or flames to detect fire, pedestrians to prevent traffic
collisions or persons to detect victims in disaster areas. Detecting
these characteristic elements is common to many of the algorithms
proposed for emergency management with computer vision. The primary
goal of many of these algorithms is to prevent or detect human harm
and therefore humans are a common characteristic element of these
emergency scenarios. The histogram of oriented gradients (HOG)
descriptor was introduced to detect pedestrians and
is one of the most used shape descriptors to
detect humans. Among the algorithms studied in this survey, HOG was
used to detect persons in disaster
areas~\cite{andriluka2010vision,soni2012classifier,soni2013victim}, to
detect pedestrians for on board vehicle cameras
\cite{suard2006pedestrian,wojek2009multi} but also to characterize
other objects such as flames, smoke and traffic
signs~\cite{barmpoutis2014smoke,wang2013spatial,zaklouta2011warning}
and as a descriptor for a tracking
algorithm~\cite{garate2009crowd} in crowd event recognition. Local binary patterns is also a well
known shape descriptor which is used to describe textures and was successfully applied to the task of smoke detection~\cite{tian2011smoke,ye2015dynamic}.

\subsubsection{Temporal features}
Most of the input data used in emergency management comes in form of
videos, which provide visual information as well as temporal
information. Here we briefly summarize the main features which exploit
the available temporal information in the data.

\minisection{Wavelets:} Flicker is an important characteristic of
smoke and flames and has been used in several systems to detect fire
and smoke in video data. Fourier analysis is the standard tool for
analysis of spatial frequency patterns. The extension of frequency
analysis to spatial-temporal signals was first studied by
Haar~\cite{haar1910theorie} and later led to the wavelet
transform~\cite{mallat1989theory}.

In \cite{toreyin2006computer} the authors study the periodic behaviour of the
boundaries of the moving fire-colored regions. The flickering flames
can be detected by finding zero crossings of the wavelet transform
coefficients which are an indicator of activity within the
region. In~\cite{xu2007automatic} the flicker of smoke is detected
computing the temporal high-frequency activity of foreground pixels by
analysing the local wavelet energy. They also use frequency
information to find smoke boundaries by detecting the decrease in high
frequency content since the smoke is semi-transparent at its
boundaries. In~\cite{calderara2008smoke} the authors present an
algorithm which uses a Discrete Wavelet Transform to analyse the
blockwise energy of the image. The variance over time of this energy
value was found to be an indicator of the presence or absence of
smoke. In~\cite{gonzalez2010wavelet} the authors perform a vertical,
horizontal and diagonal frequency analysis using the Stationary
Wavelet Transform and its inverse. A characterization of smoke
behaviour in the wavelet domain was done in~\cite{gubbi2009smoke} by
decomposing the image into wavelets and analyzing its frequency
information at various scales.

\minisection{Optical flow:} Optical flow estimates the motion of all
the pixels of a frame with respect to the previous
frame~\cite{beauchemin1995computation,lucas1981iterative}. The
objective of algorithms based on optical flow is to determine the
motion of objects in consecutive frames. Many techniques have been proposed in
the literature to perform optical flow~\cite{brox2004high,brox2011large,gautama2002phase}, however the
most widely used technique is the one proposed by Lucas-Kanade~\cite{lucas1981iterative} with some
extensions~\cite{bouguet2001pyramidal}.

Optical flow algorithms have been used in several computer vision
systems for emergency management, such as for the detection of fire
and smoke and anomalous situations on roads or in crowds. In the case
of fire and smoke detection several approaches use the pyramidal
Lucas-Kanade optical flow algorithm to determine the flow of candidate
regions which likely to contain fire or smoke and to discard regions
that do not present a high variation on the flow rate characteristic
of the fire turbulence~\cite{rinsurongkawong2012fire,yu2013real}. The
pyramidal Lucas-Kanade optical flow algorithm assumes that the objects
in the video are rigid with a Lambertian surface and negligible
changes in brightness but the drawback is that these assumptions do
not hold for fire and smoke. Addressing this problem, Kolesov et
al.~\cite{kolesov2010natural} proposed a modification to Lucas-Kanade in
which they assume brightness and mass conservation. Optical
flow has been also used to analyse the motion of crowds in order to
determine abnormal and therefore potentially dangerous
situations~\cite{andrade2006hidden,ihaddadene2008real,mehran2009abnormal}. In~\cite{sultani2010abnormal}
they use an optical flow technique to model the traffic dynamics of
the road, however instead of doing a pixel by pixel matching they
consider each individual car as a particle for the optical flow model.

\minisection{Background modeling and
  subtraction:} 
The separation of background from foreground is an important step in
many of the applications for emergency management. Background modeling
is an often applied technique for applications with static cameras
(see Section~\ref{minisec:static}) and static background. The aim of
this method is to compute a representation of the video scene with no
moving objects, a comprehensive review can be seen in~\cite{piccardi2004background}.

In the context of emergency management, the method of Stauffer and
Grimson~\cite{stauffer2000learning} has been used to segment fire,
smoke and persons from the background
\cite{andrade2006modelling,calderara2008smoke,lu2004vision,thome2008falling,truong2012fire,wang2013spatial}. Although not as popular, other background subtraction algorithms have been
proposed in the literature~\cite{collins2000system,izadi2008robust,kim2005real,li2003foreground}
and used in systems for emergency
management~\cite{ihaddadene2008real,ko2010early,liao2012slip,shieh2012falling,toreyin2006computer}. An
interesting application of background modeling was described
in~\cite{kamijo2000traffic} in the context of abnormal event detection
in emergency situations. They assume that vehicles involved in an
accident or an abnormal situation remain still for a considerable
period of time. Therefore an abnormal situation or accident can be
detected in a surveillance video by comparing consecutive background
models.

\minisection{Tracking:}  Tracking in computer vision is the
problem of locating moving objects of interest in consecutive frames
of a video. The shape of tracked objects can be represented by points, primitive
geometric shapes or the object
silhouette~\cite{yilmaz2006object}. Representing the object to be
tracked as points is particularly useful when object boundaries are
difficult to determine, for example when tracking
crowds~\cite{ihaddadene2008real} where detecting and tracking each
individual person may be infeasible. It is also possible to represent
the tracked objects as primitive objective shapes, for example
in~\cite{lu2004vision} the authors use ellipses to represent swimmers
for the detection of drowning incidents. It is also common to
represent the tracked objects as rectangles, e.g. in swimmer
detection~\cite{chen2011framework} and traffic accident
detection~\cite{kamijo2000traffic}. Finally, tracked objects can also
be represented with the object's silhouette or contour~\cite{kam2002video}.

Feature selection is crucial for tracking accuracy. The features
chosen must be characteristic of the objects to be tracked and as
unique as possible. In the context of fall detection, the authors
of~\cite{hazelhoff2008video} use skin color to detect the head of the
tracked person. Many tracking algorithms use motion as the main
feature to detect the object of interest, for example applications in
traffic emergencies~\cite{hu2006system,kamijo2000traffic}, fall
detection~\cite{jiang2013real,liao2012slip}, and drowning person
detection~\cite{lu2004vision}. Haar-like
features~\cite{viola2004robust} have been widely used in computer
vision for object detection, and have been applied to swimmer
detection~\cite{chen2011framework}. Finally, descriptors such as
Harris~\cite{harris1988combined} or HOG~\cite{dalal2005histograms} can
also be useful for crowd analysis
\cite{garate2009crowd,ihaddadene2008real}.

\subsubsection{Convolutional features}

Recently, convolutional features have led to a very good performance in analyzing visual imagery. Convolutional Neural Networks (CNN) were first proposed in 1990 by LeCun et al. \cite{lecun1990handwritten}. However, due to the lack of training data and computational power, it was not until 2012 \cite{krizhevsky2012imagenet} that these networks started to be extensively used in the computer vision community.

CNNs have also been used in the recent years for some emergency challenges, mainly fire and smoke detection in images or video and anomalous event detection in videos. For the task of smoke detection, some researchers proposed novel networks to perform the task \cite{frizzi2016convolutional,xu2017deep}. In \cite{frizzi2016convolutional} the authors propose a 9-layer convolutional network and they use the output of the last layer as a feature map of the regions containing fire, while in \cite{xu2017deep} the authors construct a CNN and use domain adaptation to alleviate the gap between their synthetic fire and smoke training data and their real test data.

Other researchers have concentrated on solutions built using existing networks \cite{lagerstrom2016image,maksymiv2016deep}. In \cite{maksymiv2016deep} the authors retrain some existing architectures, while in \cite{lagerstrom2016image} the authors use already trained architectures to give labels to the images and then classify those labels into fire or not fire. For the task of anomaly detection,  algorithms need to take temporal and spatial information into account. To handle this problem the authors in \cite{chong2017abnormal,medel2016anomaly} propose to use a Long Short-Term Memory (LSTM) to predict the evolution of the video based on previous frames. If the reconstruction error between subsequent frames and the predictions is very high it is considered to be an anomaly. In \cite{zhou2016spatial} the authors consider only the regions of the video which have changed over the last period of time and send those regions through a spatial-temporal CNN. In \cite{sabokrou2016fully} the authors pass a pixel-wise average of every pair of frames through a 4-layered fully connected network and an autoencoder trained on normal regions.

\subsection{Machine learning} 
In the vast majority of the applications in emergency management
machine learning algorithms are applied. They are used for
classification and regression to high-level semantic information, to
impose spatial coherence, as well as to estimate uncertainty of the
models. Here we briefly discuss the most used algorithms.


\subsubsection{Artificial Neural Networks}

Artificial Neural Networks (ANN) have received considerable interest due to their wide range of applications and ability to handle problems involving imprecise and complex nonlinear data. Moreover, their learning and prediction capabilities combine with impressive accuracy values for a wide range of problems. A neural network is composed of a large number of interconnected nonlinear computing elements or neurons organized in an input layer, a variable number of hidden layers and an output layer. The design of a neural network model requires three steps: selection of the statistical variables or features, selection of the number of layers (really the hidden layers) and nodes and selection of transfer function.   So, the different statistical variables or features, the number of nodes and the selected transfer function, make the difference between the papers found showing applications of the ANN to prevention and detection of emergencies using computer vision and image processing techniques.

Applications of ANNs to fire and smoke detection and prevention can be seen in papers \cite{kolesov2010natural,xu2007smoke,yu2013real,yuan2011video}. They mainly differ in the characteristics of the network. In \cite{xu2007smoke} the authors extract features of the moving target which are normalized and fed to a single-layer artificial neural network with five inputs, one hidden layer with four processing elements, and one output layer with one processing element to recognize fire smoke. In \cite{yu2013real} they propose a neural network to classify smoke features.  In \cite{kolesov2010natural} they implement an ANN fire/smoke pixel classifier. The classifier is implemented as a single-hidden-layer neural network with twenty hidden units, with a softmax non-linearity in the hidden layer. The method proposed in \cite{yuan2011video} describes a neural network classifier with a hidden layer is trained and used for discrimination of smoke and non-smoke objects. The output layer uses a sigmoid transfer function.


In the system presented in \cite{alhimale2014fallin} a neural network of perceptrons was used to classify falls against the set of predefined situations. Here, the neural network analyses the binary map image of the person and identifies which plausible situation the person is at any particular instant in time.

In general, the usage of ANNs provide a performance improvement compared to other learning methods \cite{chunyu2010smoke}. The results rely highly on the selected statistical values for training. If the selection of statistical values is not appropriate, the results on the accuracy might be lower and simpler machine learning algorithm may be preferred.

\minisection{Deep Learning}
ANN algorithms that consist of more than one hidden layer are usually called \emph{deep} learning algorithms. Adding more layers to the network increases the complexity of the algorithm making it more suitable to learn more complex tasks. However, increasing the layers of the network comes at a higher computational cost and a higher probability to overfit or converge to a local minima. Nevertheless, the advances in computers and the increase in training data has facilitated the research towards deeper models which have proven to improve the results.

In \cite{chunyu2010smoke} the authors propose a multi-layered ANN for smoke features classification. In the works \cite{alhimale2014fallin,foroughi2008falling,juang2007falling,sixsmith2004smart} we find applications of ANNs to prevent and detect single-person emergencies. A neural network is used to classify falls in \cite{sixsmith2004smart} using vertical velocity estimates derived either directly from the sensor data or from the tracker. The network described in \cite{juang1998neural}, with four hidden layers, is used in the classifier design to do posture classification that can be used to determine if a person has fallen. In \cite{foroughi2008falling} the authors monitor human activities using a multilayer perceptron with focus on detecting three types of fall.

\subsubsection{Support Vector Machines}
Support vector machines (SVMs) are a set of supervised machine learning algorithms based on statistical learning and risk minimization which are related to classification and regression problems. Given a set of labeled samples a trained SVM creates an hyperplane or a set of them, in a high dimensional space, that optimally separates the samples between classes maximizing the separation margin among classes. Many computer vision systems related to emergency situations can be considered as a binary classification problems: fire or no fire, fall or no fall, crash or no crash. To solve this classification problem many researches make use of SVMs \cite{mirmahboub2013automatic,suard2006pedestrian,tian2011smoke,truong2012fire} based on the classification performance that they can reach, their effectiveness to work on high dimensional spaces and their ability to achieve good performance with few data.

In more complex systems where it is difficult to predict the type of
emergencies that may occur, for example in systems that supervise
crowded areas or traffic, algorithms that consider abnormal behaviours
as potential emergencies have been proposed
\cite{piciarelli2008trajectory,wang2012histograms}. A difficulty of
these kind of systems is that they can only be trained with samples of
one class (normal scenes). For these kind of problems one-class SVMs,
which find a region of a high dimensional space that contains the
majority of the samples, are well suited.

\subsubsection{Hidden Markov Models}

Hidden Markov Models (HMMs), which are especially known for their
application in temporal pattern recognition such as speech,
handwriting, gesture and action
recognition~\cite{yamato1992recognition}, are a type of stochastic
state transition model \cite{rabiner1986introduction}. HMMs make it
possible to deal with time-sequence data and can provide time-scale
invariability in recognition. Therefore it is one of the preferred
algorithms to analyze emergency situations subject to time
variability. 

To detect smoke regions from video clips, a dynamic texture descriptor is proposed in \cite{ye2015dynamic}. The authors propose a texture descriptor with Surfacelet transformation (\cite{lu2007surfacelets}) and Hidden Markov Tree model (HMT). Examples of use of Markov models to study human-caused emergencies that affect a single person can be seen in the following works \cite{chen2010hidden,eng2003automatic,jiang2013real,thome2008falling,toreyin2006falling},  covering fall detection and detection behavior of a swimmer and drownings. Hidden Markov Models have been applied on monitoring emergency situations in crowds by learning patterns of normal crowd behaviour in order to identify unusual or emergency events, as can be seen in \cite{andrade2006hidden,andrade2006detection}. In these two works Andrade and his team presents similar algorithms in order to detect emergency events in crowded scenarios, and based on optical flow to extract information about the crowd behaviour and the Hidden markov Models to detect abnormal events in the crowd.

T\"oreyin and his team in \cite{toreyin2006falling} use  Hidden Markov Models to recognize possible human motions in video. Also they use audio track with a three-state HMM to distinguish a person sitting on a floor from a person stumbling and falling. In \cite{thome2008falling}  the motion analysis is performed by means of a layered hidden markov model (LHMM). A fall detection system based on motion and posture analysis with surveillance video is presented in \cite{jiang2013real}.  The authors propose a context-based fall detection system by analyzing human motion and posture using a discrete hidden Markov model. The problem of recognition of several specific behaviours of a swimmer (including backstroke, breaststroke, freestyle and butterfly, and abnormal behaviour)  in a swimming pool using Hidden Markov Models is addressed in \cite{chen2010hidden}.  The algorithm presented in \cite{eng2003automatic} detects water crises within highly dynamic aquatic environments where partial occlusions are handled using a Markov Random Field framework.

Hidden Markov Models have been applied on monitoring emergency situations in crowds by learning patterns of normal crowd behavior in order to identify unusual or emergency events, as can be seen in \cite{andrade2006hidden,andrade2006detection}. In these two works Andrade and his team presents similar algorithms in order to detect emergency events in crowded scenarios, and based on optical flow to extract information about the crowd behavior and the Hidden Markov Models to detect abnormal events in the crowd.

\subsubsection{Fuzzy logic}
Fuzzy sets theory and fuzzy logic are powerful frameworks for performing automated reasoning, and provide a principled approach to address the inherent uncertainty related to modeling; an aspect which relevant in many emergency management application. Fuzzy set theory, originally introduced by Lofti A. Zadeh in 1965 \cite{Zad65}, is an extension of classical set theory where to each element $x$ from a set $A$ is associated with a value between $0$ and $1$, representing its degree of belonging to the set $A$. One important branch of the fuzzy set theory is fuzzy logic \cite{Zad73}. An inference engine operators on rules that are structured in an IF-THEN rules, are presented. Rules are constructed from linguistic variables. A fuzzy approach has the advantage that the rules and linguistic variables are understandable and simplify addition, removal, and modification of the system's knowledge.

In the papers \cite{anderson2009falling}  and \cite{thome2008falling} the authors proposed a multi-camera video-based methods for monitoring human activities, with a particular interest to the problem of fall detection. They use methods and techniques from fuzzy logic and fuzzy sets theory.  Thome et al. \cite{thome2008falling} apply a fuzzy logic fusion process that merges features  of each camera to provide a multi-view pose classifier efficient in unspecified conditions (viewpoints).  In \cite{anderson2009falling}, they construct a three-dimensional representation of the human from silhouettes  acquired from multiple cameras monitoring the same scene, called voxel person. Fuzzy logic is used to determine the membership degree of the person to a pre-determined number of states at each image. A method, based on temporal fuzzy inference curves representing the states of the voxel person, is presented for generating a significantly smaller number of rich linguistic summaries of the humans state over time with the aim to detect of fall detection.

Another problem addressed with methods and techniques from theory of fuzzy sets is fire prevention by smoke detection. It has been observed that the bottom part of a smoke region is less mobile than top part over time; this is called swaying motion of smoke. In \cite{wang2014smoke} a method to detect early smoke in video is presented, using swaying motion and a diffusion feature.  In order to realized data classification based on information fusion,  Choquet fuzzy integral (see \cite{grabisch2000fuzzy}) was adopted to extract dynamic regions from video frames, and then, a swaying identification algorithm based on centroid calculation was used to distinguish candidate smoke region from other dynamic regions. Finally, Truong et al. \cite{truong2012fire} presented a four-stage smoke detection algorithm, where they use a fuzzy c-means algorithm method to cluster candidate smoke regions from the moving regions.

\subsection{Algorithmic trends in emergency management}
\label{sec:trends}
\begin{figure*}
\begin{center}
\begin{tabular}{ccc}
\subfloat[Algorithms used for fire emergency management.]{\includegraphics[width = 0.3\linewidth]{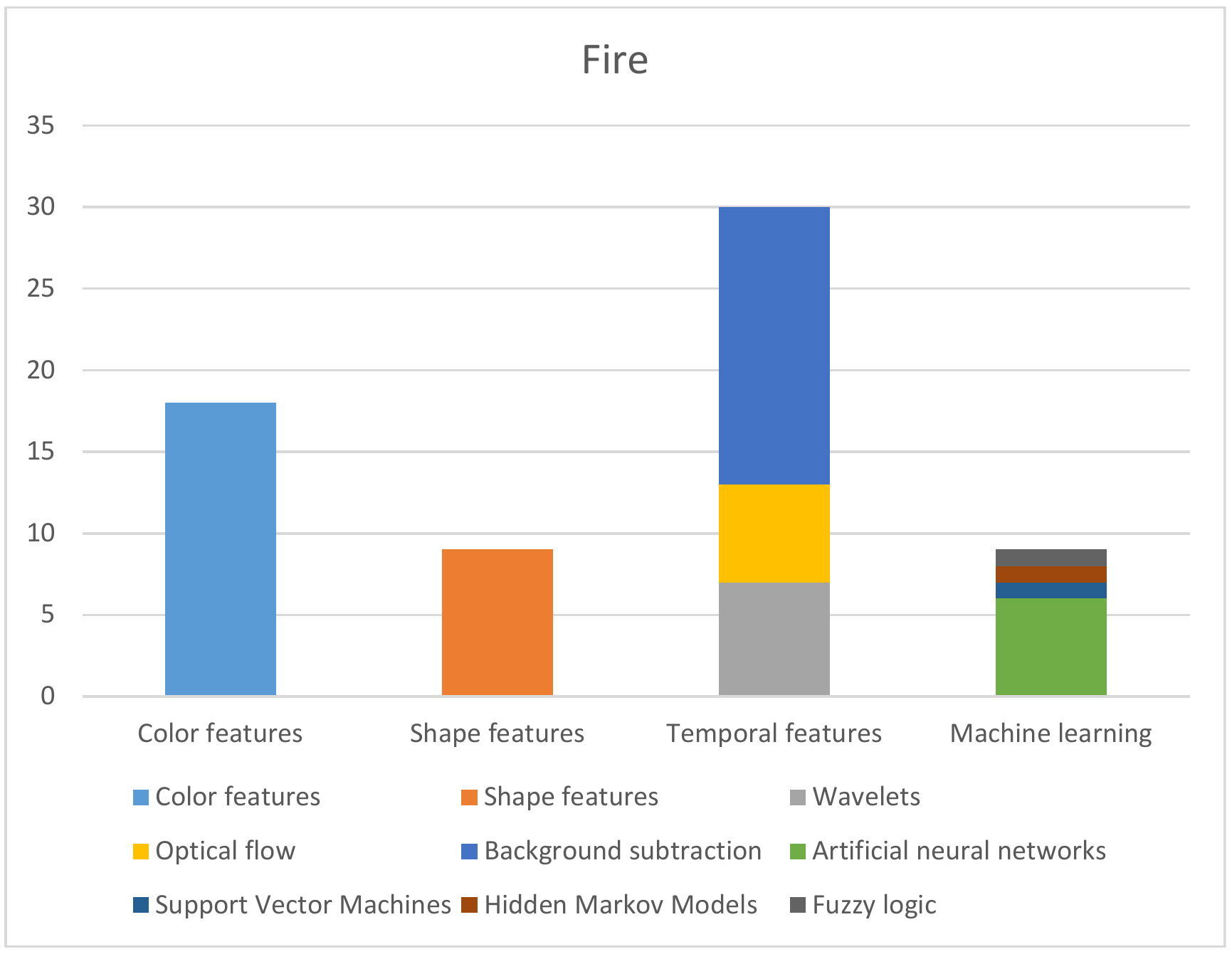}}&
\subfloat[Algorithms used for road accidents emergency management.]{\includegraphics[width = 0.3\linewidth]{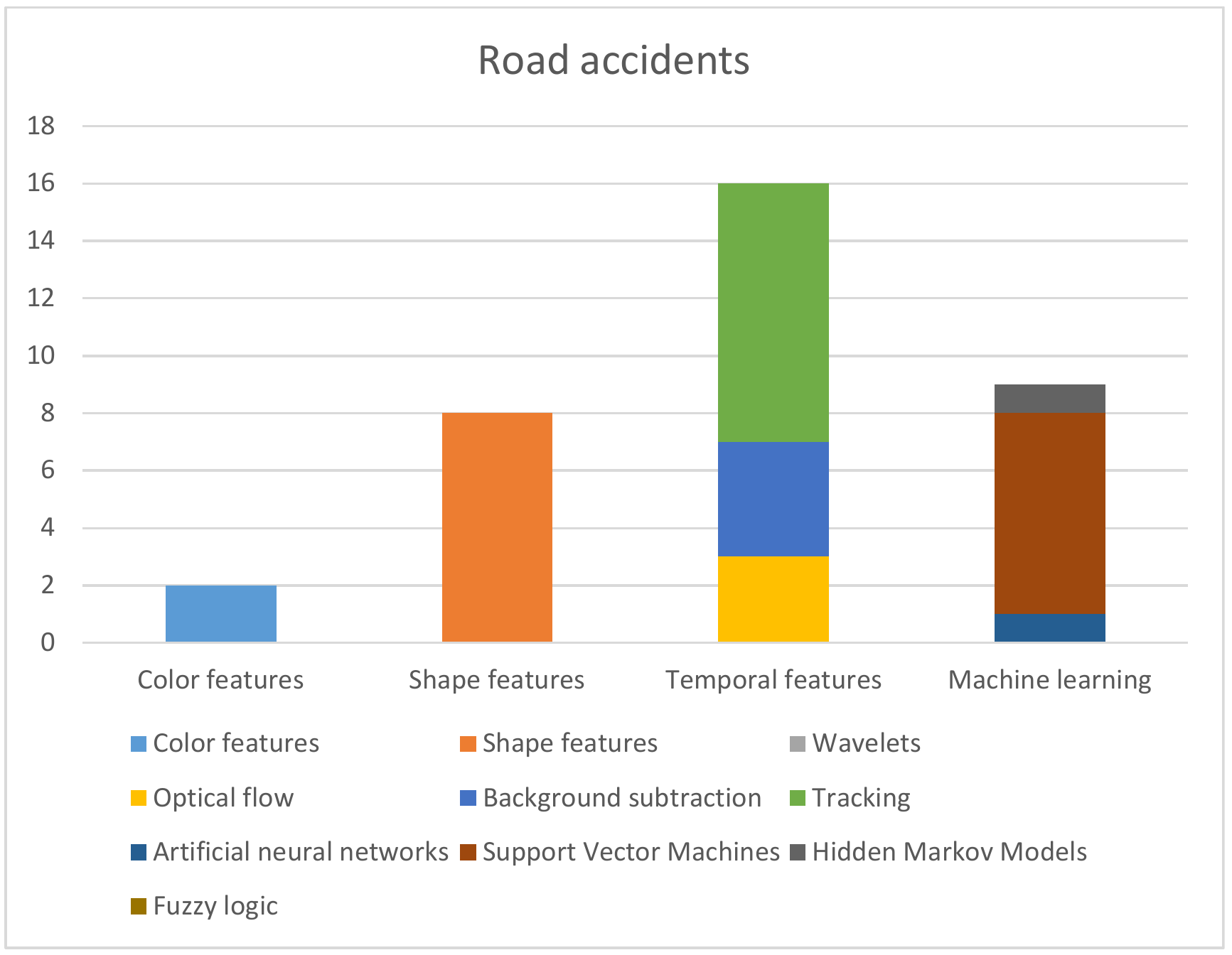}} &
\subfloat[Algorithms used for human fall emergency management.]{\includegraphics[width = 0.3\linewidth]{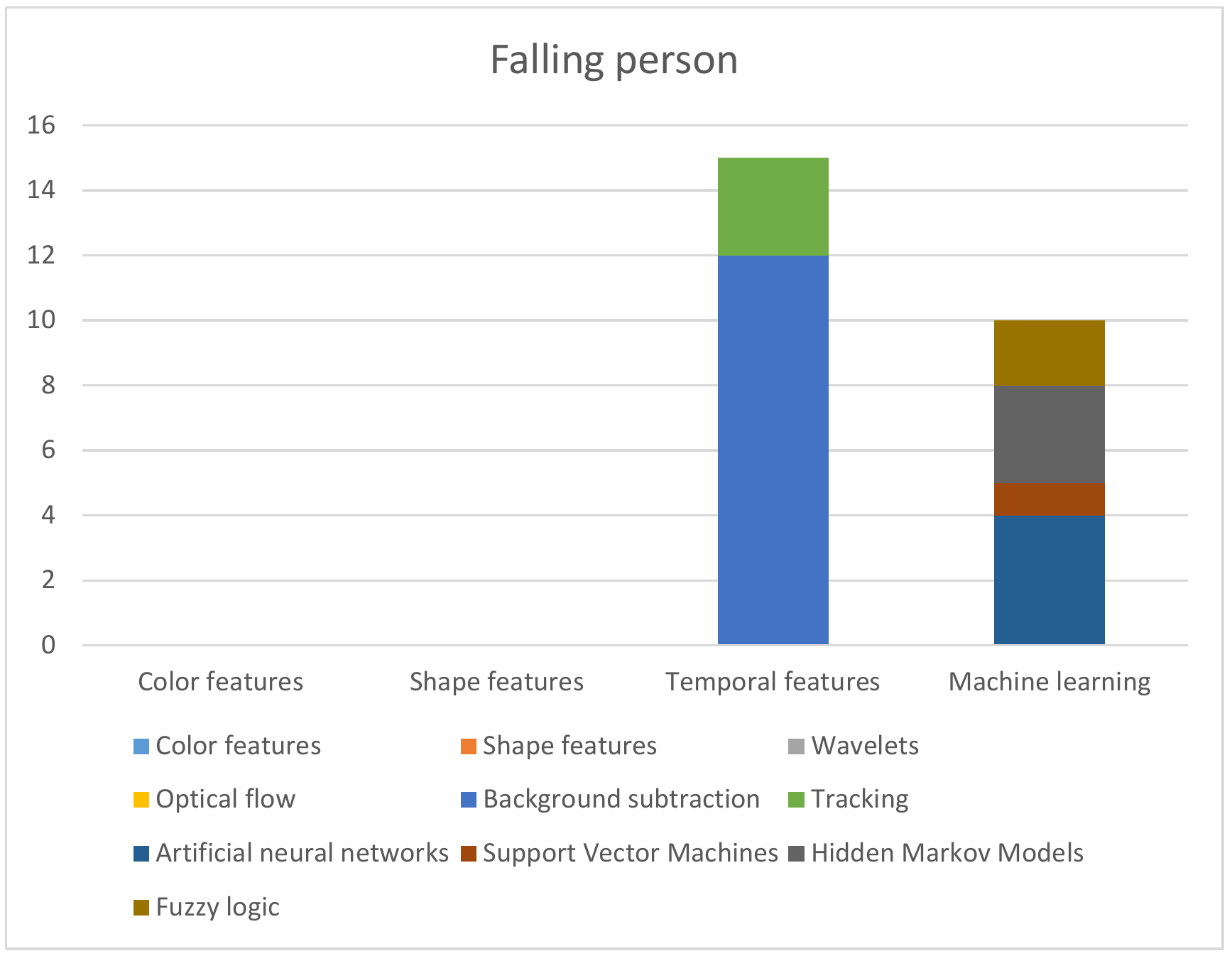}}
\end{tabular}
\caption{Overview of the algorithms that have been used to manage the three most studied emergency scenarios.}
\label{fig:graphs}
\end{center}
\end{figure*}

To analyze the usage of the discussed algorithms across emergency
types, we provide a more in depth analysis of the three most studied
emergencies: fire, road accidents, and human fall. In
Figure~\ref{fig:graphs}(a-c) we break down the algorithms that have
been used for these three types of emergencies. As can be seen in the
figures, for the three types of emergencies, the most commonly applied
algorithms are the ones that use temporal features. For fire and human
fall management background subtraction is the most popular algorithm
since most times the most discriminative feature of the objects of
interest in this scenarios move on top of a static background. In the
case of road accident management it is common to use non-static
cameras (such as cameras on board a vehicle) and therefore the
background is not static and background subtraction algorithms
are not useful. However, temporal features are used to perform optical
flow or tracking in order to determine the trajectory of the objects
in the road or close by. For fire color features are very distinctive
and therefore more researched.


\section{Discussion}
There are many video monitoring systems whose aim is to monitor
various types of emergency scenarios. Since continuous human vigilance
of these recordings is infeasible, most of this data is used \emph{a
  posteriori}. Although this information is very useful for the later
understanding of the event or as an incriminatory proof, many
researchers study the possibility of exploiting these huge amounts of
data and analysing them in real time with the hope of preventing some
of these emergencies or to facilitate a faster or more efficient
response. In this survey we reviewed and analyzed the progress of
computer vision covering many types of emergencies with the goal of
unveiling the overlap across emergency research in computer vision,
especially in the hardware and algorithms developed, making it easier
to researchers to find relevant works.

In order to restrict the scope of this survey we first gave a description
of what we consider emergency situations. Taking into account the
description provided, we restrict ourselves to emergency scenarios
that have been studied in the context of computer vision. Taking all
these considerations into account we identified eleven different
emergency scenarios, studied in computer vision, which we classified into natural emergencies and
emergencies caused by humans.  After a thorough analysis of the
state-of-the-art algorithms on these topics, we inferred that there
are four main axes that have to be taken into account when creating
computer vision systems to help in emergency scenarios: the type of emergency that is going to be addressed, the monitoring
objective, the hardware and sensors needed, and the algorithms
used.

When talking about the monitoring objective, we have seen that prevention through computer vision techniques can only be done when there is visual evidence of risk factors such as for fire detection using smoke features, car accidents by detecting pedestrians, driver drowsiness and unusual behaviours, and finally fall prevention falls by detecting slips. The most common objective is emergency detection, which has been studied in all the emergencies studied in this survey. It is also common to use computer vision to detect humans in disaster areas to assist emergency responders. Finally, computer vision has not played a role from the emergency understanding point of view.

  From all the acquisition methods studied in Section \ref{sec:acquisition} we have concluded that the most common sensors are monocular RGB or grayscale cameras -- mainly for their low cost -- and that they are normally situated in fixed locations. However, we see a trend in the use of UAVs which are becoming cheaper and provide great flexibility through their mobility.

Finally, as discussed in Section \ref{sec:trends}, it is very difficult to draw conclusions on algorithmic trends for every emergency since some of them have little impact in computer vision. However, there is a clear tendency towards using color and temporal features for fire emergencies, a clear preference for temporal features in road accidents, and tracking algorithms and background subtraction techniques are the most common when detecting falling person events.

This up-to-date survey will be important for helping readers obtain a global view. At the same time they can find novel ideas of how computer vision can be helpful to prevent, respond and recover from emergency scenarios, and learn about some required hardware and relevant algorithms that are developed to solve these problems.

It is extremely difficult to compare different algorithms proposed in the literature for solving similar problems due to the lack of unified benchmarks. From all the papers studied during this review, over 60\% used non-publicly available datasets. In order to facilitate a common benchmark to compare algorithms, in Table \ref{tab:datasets} we report some of the most important publicly available datasets used among the emergencies studied in this survey. Note that some pedestrian datasets are shared among emergencies since detecting humans is considered useful in more than one emergency.

\begin{table*}[h!]
\begin{center}
\begin{tabular}{p{1.65cm}p{3.5cm}p{5cm}p{5cm}}
\toprule
{\bf Type of emergency} & {\bf Public datasets} & {\bf General comments} & {\bf Url} \\
\toprule
\multirow{5}{*}{Fire} &  \href{http://imagelab.ing.unimore.it/visor/video\_videosInCategory.asp?idcategory=-1}{VISOR smoke dataset} & Videos with evidence of smoke. & \url{http://imagelab.ing.unimore.it/visor}\\
& \href{http://signal.ee.bilkent.edu.tr/VisiFire/}{Fire dataset from the Bilkent University}& Videos with evidence of smoke and fire. & \url{http://signal.ee.bilkent.edu.tr/VisiFire/}\\
& \href{http://www.mesh-ip.eu/?Page=Project}{MESH database of news content} & Instances of catastrophe related videos from the Deutsche Welle broadcaster with several news related to fires in a variety of conditions. & \url{http://www.mesh-ip.eu/} \\
& \href{http://fire.nist.gov/fastdata}{FASTData} & Collection of fire images and videos. & \url{http://fire.nist.gov/fastdata}\\
& \href{http://signal.ee.bilkent.edu.tr/VisiFire/Demo/SampleClips.html}{Fire and smoke dataset} & Clips containing evidence of fire and smoke in several scenarios. & \url{http://signal.ee.bilkent.edu.tr/VisiFire/Demo} \\
                                                 
\midrule
\multirow{3}{*}{Fall detection} & \href{http://www.derektanderson.com/fallrecognition/datasets.html}{CIRL fall detection dataset} & Indoor videos for action recognition and fall detection, not available for comercial purposes. & \url{http://www.derektanderson.com/fallrecognition/datasets.html}\\
& \href{http://foe.mmu.edu.my/digitalhome/FallVideo.zip}{MMU fall detection dataset} & 20 indoor videos including 38 normal activities and 29 different falls  & \url{http://foe.mmu.edu.my/digitalhome/FallVideo.zip}\\
& \href{http://www.iro.umontreal.ca/~labimage/Dataset/}{Multiple camera fall dataset} & 24 videos of falls and fall confounding situations recorded with 8 cameras in different angles. & \url{http://www.iro.umontreal.ca/~labimage/Dataset/} \\

\midrule
\multirow{2}{*}{Injured civilians} & \href{https://research.csiro.au/data61/automap-datasets-and-code/}{NICTA dataset} & Contains a total of 25551 unique pedestrians. & \url{https://research.csiro.au/data61/automap-datasets-and-code}\\
& \href{http://cbcl.mit.edu/software-datasets/PedestrianData.html}{MIT pedestrian dataset} & 924 images of 64x128 containing pedestrians. & \url{http://cbcl.mit.edu/software-datasets/PedestrianData.html}\\

\midrule
\multirow{7}{*}{Road accident} & \href{http://cbcl.mit.edu/software-datasets/PedestrianData.html}{MIT pedestrian dataset} & 924 images of 64x128 containing pedestrians. & \url{http://cbcl.mit.edu/software-datasets/PedestrianData.html} \\
& \href{http://pascal.inrialpes.fr/data/human/}{INRIA pedestrian dataset} & Images of pedestrians and negative images. & \url{http://pascal.inrialpes.fr/data/human/}\\
& \href{http://groups.inf.ed.ac.uk/vision/CAVIAR/CAVIARDATA1/}{CAVIAR dataset} & Sequences containing pedestrians, also suitable for action recognition. & \url{http://groups.inf.ed.ac.uk/vision/CAVIAR}\\
& \href{http://ieeexplore.ieee.org/document/4633642/}{Trajectory based anomalous event detection} & Synthetic and real-world data. & \url{http://ieeexplore.ieee.org/document/4633642/}\\
& \href{http://benchmark.ini.rub.de/?section=gtsdb&subsection=dataset}{The German Traffic Sign detection benchmark} & 900 images of roads with traffic sign ground truth. & \url{http://benchmark.ini.rub.de}\\
& \href{http://vcipl-okstate.org/pbvs/bench/}{Thermal pedestrian database} & 10 sequences containing pedestrians recorded with a thermal camera. & \url{http://vcipl-okstate.org/pbvs/bench/}\\
& \href{https://www.mpi-inf.mpg.de/departments/computer-vision-and-multimodal-computing/research/people-detection-pose-estimation-and-tracking/multi-cue-onboard-pedestrian-detection/}{TUD Brussels and TUD Paris, Multi-Cue Onboard Pedestrian Detection} & Images of pedestrians taken by on-board cameras. & \url{ttps://www.mpi-inf.mpg.de/departments/computer-vision-and-multimodal-computing/research/people-detection-pose-estimation-and-tracking} \\

\midrule
\multirow{5}{*}{Crowd related} & \href{http://citeseerx.ist.psu.edu/viewdoc/summary?doi=10.1.1.59.8399}{Simulation of Crowd Problems for Computer Vision} & Approach for generating video evidence of dangerous situations in crowded scenes. & \url{https://www.icsa.inf.ed.ac.uk/publications/online/0493.pdf} \\
& \href{http://www.cvg.reading.ac.uk/PETS2009/a.html}{PETS dataset} & Sequences containing different crowd activities. & \url{http://www.cvg.reading.ac.uk/PETS2009/a.html} \\
& \href{http://mha.cs.umn.edu/Movies/Crowd-Activity-All.avi}{Unusual crowd activity dataset} & Clip with unusual crowd activities. & \url{http://mha.cs.umn.edu/Movies/Crowd-Activity-All.avi}\\
& \href{http://www.svcl.ucsd.edu/projects/anomaly/dataset.htm}{USCD anomaly detection dataset} & Clips of the street with stationary camera. & \url{http://www.svcl.ucsd.edu/projects/anomaly/dataset.htm}\\
& \href{http://mha.cs.umn.edu/proj_recognition.shtml#crowd_count}{UMN monitoring human activity dataset} & Videos of crowded scenes. &  \url{http://mha.cs.umn.edu/proj_recognition.shtml#crowd_count}\\

\midrule

\end{tabular}
\end{center}
\caption{Recompilation of the publicly available datasets used in the papers reviewed in this manuscript. Only about 40\% of the papers reviewed used datasets publicly available.}
\label{tab:datasets}
\end{table*}

In general, we have seen a tendency to create more complex systems
that combine different algorithms or algorithms of a higher
computational complexity which in many cases could not have been
feasible with the previous generations of computers. Among these
computationally complex algorithms, the emerging techniques that stand
out are those based on Convolutional Neural Networks (CNN) and Deep
Learning~\cite{gope2016deep,kang2016deep,zhang2016deep}. The field of
computer vision is experiencing rapid change since the rediscovery
of CNNs. These algorithms have already been introduced in some
emergency scenarios studied in this survey and we believe that the
impact of CNNs will grow over the years in the field of emergency
management with computer vision. Moreover, with the decreasing cost
of visual acquisition devices such as infrared cameras or stereo
cameras, there is a trend of combining different information sources
which can help to increase accuracy rates and decrease false
alarms. We also believe that, with the help of this survey and further
study of the overlap that exists between the algorithms developed to
detect and study different types of emergency situations, it will
be possible in the future to create system that can cover more than
one type of emergency.


\label{sec:discussion}

%
%


\bibliographystyle{spmpsci}      
\bibliography{refs}   


\end{document}